\documentclass[letterpaper]{article} 
\usepackage[preprint]{aaai2027}  
\usepackage[hyphens]{url}  
\usepackage{graphicx} 
\usepackage{natbib}  
\usepackage{caption} 

\usepackage{algorithm}

\usepackage[most]{tcolorbox}

\usepackage{newfloat}
\usepackage{listings}
\DeclareCaptionStyle{ruled}{labelfont=normalfont,labelsep=colon,strut=off} 
\floatstyle{ruled}
\newfloat{listing}{tb}{lst}{}
\floatname{listing}{Listing}

\usepackage{booktabs}       
\usepackage{amsfonts}       
\usepackage{nicefrac}       
\usepackage{microtype}      
\usepackage{xcolor}         
\usepackage{graphicx}
\usepackage{amsmath}
\usepackage{amssymb}
\usepackage{array}
\usepackage{multirow}
\usepackage{color}
\usepackage{colortbl}
\usepackage{framed}
\usepackage{bm}
\usepackage{xspace}
\usepackage{enumitem}
\usepackage{xparse}
\usepackage{algorithm}
\usepackage{listings}
\usepackage{url}
\usepackage{mathtools}
\usepackage{pifont}
\usepackage{mathtools}
\usepackage{lipsum}
\usepackage{adjustbox}
\usepackage{amssymb}
\usepackage{multirow}
\usepackage{graphicx} 
\usepackage[table]{xcolor}
\usepackage{algpseudocode}
\usepackage{soul} 
\usepackage{mathtools}
\usepackage{xcolor}

\newcommand{\cmark}{\ding{51}}%

\definecolor{Gray}{gray}{0.2}
\definecolor{lightgray}{gray}{0.92}
\definecolor{blond}{rgb}{0.98, 0.94, 0.75}
\definecolor{TitleColor}{gray}{0.95}
\definecolor{LightCyan}{rgb}{0.88,0.95,1}
\definecolor{OurColor}{RGB}{255,192,156}
\colorlet{OurColor}{OurColor!30}

\def \ie {\emph{i.e.}}
\def \eg {\emph{e.g.}}

\newcommand{\tit}[1]{\smallbreak\noindent\textbf{#1.}}
\newcommand{\tinytit}[1]{\noindent\textbf{#1.}}

\definecolor{MyGreen}{RGB}{34,139,34}
\newcommand{\inc}[1]{\textcolor{MyGreen}{\textbf{\footnotesize$\Delta$+#1}}}

\newcommand{\oursl}{Look Twice\xspace}
\newcommand{\ours}{LoT\xspace}

\definecolor{myPink}{RGB}{247, 219, 231}  
\definecolor{myYellow}{RGB}{255, 242, 204}    
\definecolor{myPurple}{RGB}{235,225,250}    

\definecolor{systempromptbg}{RGB}{235,247,250}
\definecolor{systempromptframe}{RGB}{92,123,132}

\definecolor{usertemplatebg}{HTML}{EEF6E9}
\definecolor{usertemplateframe}{RGB}{105,125,96}

\definecolor{promptbg}{RGB}{253,241,246}
\definecolor{promptframe}{RGB}{133,102,116}
\definecolor{prompttext}{RGB}{38,38,38}

\tcbset{
    promptstyle/.style={
        enhanced,
        breakable,
        boxrule=0.5pt,
        leftrule=1.5mm,
        arc=1mm,
        outer arc=1mm,
        left=2mm,
        right=2mm,
        top=1.5mm,
        bottom=1.5mm,
        before skip=6pt,
        after skip=6pt,
        fonttitle=\bfseries\footnotesize,
        fontupper=\footnotesize\ttfamily,
        coltitle=black
    }
}

\newtcolorbox{system_prompt}[1][]{
    promptstyle,
    colback=systempromptbg,
    colframe=systempromptframe,
    coltext=black,
    title={System Prompt},
    #1
}

\newtcolorbox{user_template}[1][]{
    promptstyle,
    colback=usertemplatebg,
    colframe=usertemplateframe,
    coltext=black,
    title={User Template},
    #1
}

\newtcolorbox{promptbox3}[1][]{
    promptstyle,
    colback=promptbg,
    colframe=promptframe,
    coltext=prompttext,
    title={Prompt},
    #1
}

\newcommand{\hlpink}[1]{\sethlcolor{myPink}\hl{#1}}
\newcommand{\hlyellow}[1]{\sethlcolor{myYellow}\hl{#1}}

\newcommand{\hlours}[1]{\sethlcolor{OurColor}\hl{#1}}
\newlength{\origtextfloatsep}
\title{Look Twice: Training-Free Evidence Highlighting for\\Knowledge-based Visual Question Answering} 
\author{
    Marco Morini,
    Sara Sarto,
    Marcella Cornia,
    Lorenzo Baraldi,
    Rita Cucchiara
}
\affiliations{
    University of Modena and Reggio Emilia, Italy\\
    \{\texttt{name.surname}\}@unimore.it
    }

\begin{document}

\maketitle

\begin{abstract}
Knowledge-based Visual Question Answering (KB-VQA) requires Multimodal Large Language Models (MLLMs) to identify and combine fine-grained visual cues with retrieved textual evidence. However, retrieval often introduces noisy and partially relevant content, while images contain distracting visual regions, causing pretrained MLLMs to overlook the evidence that actually supports the answer.
To address this, we introduce \oursl (\ours), a training-free inference-time framework that turns the model's own internal attention into an explicit multimodal evidence-selection mechanism. \ours first leverages the model's internal attention patterns to identify query-relevant image regions and textual sentences, filters attention sinks and distracting content, and reformulates the input to explicitly highlight the selected evidence before answer generation. The method requires no parameter updates, auxiliary models, or architectural modifications.
Across four KB-VQA benchmarks and ten off-the-shelf MLLMs ranging from 2B to 38B parameters, \ours improves every evaluated backbone, with average gains of up to $+12.5$ accuracy points. It also provides further gains when combined with established context-refinement strategies, yielding additional improvements over already refined inputs. These results establish \ours as a general and effective mechanism for enabling pretrained MLLMs to exploit available multimodal evidence more accurately. Source code is publicly available at \url{https://aimagelab.github.io/LoT/}.
\end{abstract}

\section{Introduction}
\label{sec:intro}

Multimodal Large Language Models (MLLMs)~\cite{caffagni2024r,liu2024improved} integrate LLMs~\cite{zhao2023survey} with visual encoders~\cite{radford2021learning,zhai2023sigmoid,tschannen2025siglip} to jointly process images and text through a unified generative interface, enabling applications such as visual dialogue and open-ended question answering.
\begin{figure}[t]
\vspace{-0.25cm}
    \centering
    \includegraphics[width=\linewidth]{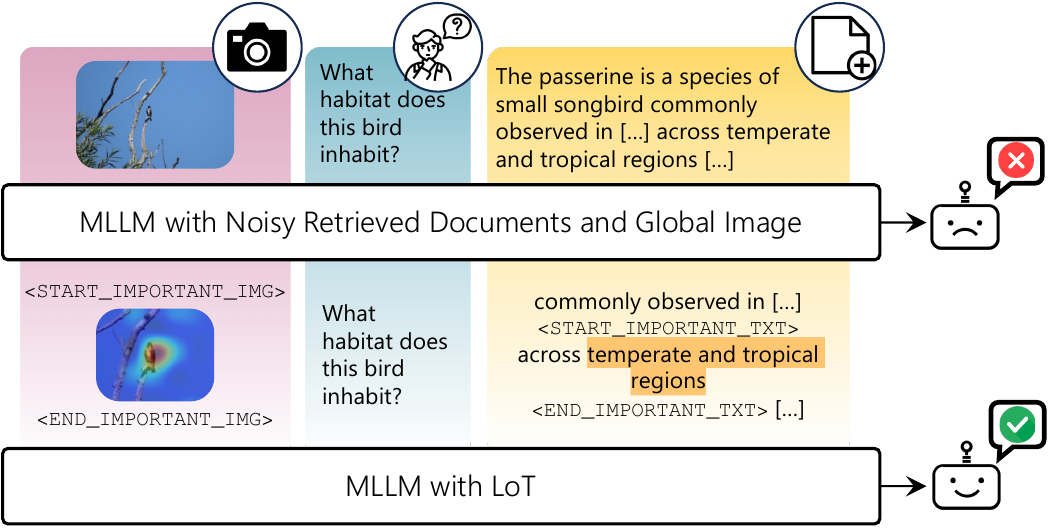}
    \vspace{-0.6cm}
    \caption{Given an image, a question, and retrieved documents, an MLLM may overlook relevant multimodal evidence and answer incorrectly. \ours uses the model's internal attention to identify relevant visual and textual content, then highlights the multimodal evidence to improve answer generation. 
    }
    \vspace{-0.5cm}
    \label{fig:first}
\end{figure}
A particularly challenging scenario arises when answering a question that requires information that is not directly observable in the image. Many real-world queries depend on external knowledge, domain-specific facts, or contextual information that can not be reliably stored in the model parameters. Knowledge-based Visual Question Answering (KB-VQA) addresses this challenge by augmenting the model input with passages retrieved from external knowledge sources~\cite{mensink2023encyclopedic,chen2023can}. In this setting, the model must reason jointly over the visual input and the retrieved textual elements to generate the final answer.

The key challenge, however, is not merely retrieving more information, but identifying and effectively exploiting the evidence that supports the answer.
Retrieved contexts often contain noisy, redundant, or only partially relevant passages~\cite{caffagni2024wiki}, which can distract the model when directly appended to the prompt.
Existing approaches address this issue through increasingly sophisticated retrieval-augmented pipelines, auxiliary reasoning modules~\cite{cocchi2025augmenting,compagnoni2025reag}, passage filtering~\cite{hong2025knowledge,ye2026qkvqa}, or re-ranking mechanisms~\cite{yan2024echosight,yang2025omgm}.
While effective, these approaches typically rely on additional training, multi-stage processing, or task-specific components, increasing computational overhead and system complexity. 

Evidence selection is equally challenging in the visual modality. Many KB-VQA questions rely on evidence localized in specific image regions, yet MLLMs often rely on coarse visual cues rather than precise spatial grounding~\cite{bi2025unveiling,deng2025words,wang2025towards}. 
This issue can be further exacerbated by \emph{attention sinks} in large Transformer architectures~\cite{attention_sinks_2025,xiao2024efficient}, where semantically uninformative tokens or patches receive disproportionate attention, thereby weakening evidence localization.
These failures reveal a common bottleneck: irrelevant passages and distracting visual regions can dominate generation despite rich multimodal inputs. KB-VQA therefore requires not only better retrieval, but also inference-time mechanisms that prioritize relevant evidence.

In this work, we investigate whether the internal attention dynamics of pretrained MLLMs can be leveraged to identify query-relevant multimodal evidence. 
Building on this, we introduce \textbf{\oursl\ (\ours)}, a \emph{training-free} inference-time framework that improves multimodal evidence selection by explicitly highlighting query-relevant cues in both retrieved text and the input image. The key idea is to let the model \emph{look twice} at the input. \ours first leverages the model's internal attention patterns to estimate the relevance of visual regions and textual elements to the query. These signals are then used to reformulate the multimodal input by highlighting the selected evidence while filtering out irrelevant or distracting content before final answer generation (Fig.~\ref{fig:first}). 

Specifically, for the visual input, \ours identifies the target entity in the question and aggregates the corresponding attention over visual tokens to obtain a query-specific relevance map. A filtering step further reduces spurious activations associated with attention sinks. For the textual input, \ours uses attention signals to identify the most informative sentences in the retrieved context. The resulting visual and textual evidence is highlighted, while irrelevant content is discarded, yielding a refined multimodal input for answer generation. Unlike existing KB-VQA methods that refine either modality through external modules, predefined regions, bounding boxes, or separately trained selectors, \ours jointly identifies and highlights evidence from both modalities using only the MLLM’s internal attention signals.

We evaluate \ours{} across four KB-VQA benchmarks~\cite{chen2023can,mensink2023encyclopedic,lerner2022viquae,hu2023open} and ten off-the-shelf MLLMs ranging from 2B to 38B parameters. In the zero-shot setting, \ours{} improves every evaluated backbone, with average gains from $+3.1$ to $+12.5$ accuracy points. Its adaptable design also enables straightforward integration into existing KB-VQA inference pipelines~\cite{wei2022chain,li2026qwen3,compagnoni2025reag,yan2024echosight}, where it provides further gains without modifying or retraining their components. Together, these results establish \ours{} as an effective and broadly applicable mechanism for improving multimodal evidence utilization across models, datasets, and inference settings.

\tinytit{Contributions} Our contributions are summarized as follows:
\begin{itemize}[noitemsep, topsep=0pt]
\item We introduce \oursl (\ours), a \emph{training-free} inference-time framework for KB-VQA that improves how pretrained MLLMs exploit multimodal evidence from retrieved text and input images.

\item We show that internal attention signals can be leveraged to jointly identify query-relevant visual and textual evidence, enabling \ours to highlight useful cues and suppress irrelevant content before answer generation.

\item We conduct an extensive evaluation across four KB-VQA benchmarks and ten MLLMs spanning 2B--38B parameters. \ours{} improves every evaluated backbone, achieving average gains up to $+12.5$, and delivers further improvements when integrated into diverse inference pipelines.

\end{itemize}

\section{Related Work}
\label{sec:related}

\tinytit{Knowledge-Based VQA} Knowledge-Based Visual Question Answering (KB-VQA) requires models to combine visual understanding with knowledge beyond what is directly observable in the image.
While early benchmarks~\cite{marino2019ok,schwenk2022okvqa} introduced this challenge, recent datasets such as Encyclopedic-VQA~\cite{mensink2023encyclopedic} and InfoSeek~\cite{chen2023can} require fine-grained reasoning over large-scale knowledge sources. To address these, recent approaches rely on sophisticated retrieval-augmented pipelines, additional supervision, or task-specific training to better integrate external evidence~\cite{caffagni2024wiki,yan2024echosight,yuan2025mkg,hong2025knowledge,cocchi2025augmenting,compagnoni2025reag,yang2025omgm}. In contrast, we investigate how pretrained MLLMs exploit retrieved textual context and visual evidence directly at inference time. Our training-free evidence highlighting strategy improves multimodal evidence selection without additional supervision, parameter updates, or architectural modifications.

\vspace{-0.04cm}
\tit{Attention Dynamics in Vision-Language Models}
Prior work has extensively analyzed how Transformer models distribute attention, revealing structured patterns involving sparsity, head specialization, and redundancy~\cite{clark2019attention,voita2019analyzing,kobayashi2020attention,michel2019sixteen,cordonnier2020relationship}.
More recent work highlights the phenomenon of \textit{attention sinks}, where a small subset of tokens or image regions captures a disproportionate amount of attention while other potentially relevant elements are ignored. In LLMs, such sinks often correspond to special or punctuation tokens and can hinder effective long-context reasoning~\cite{attention_sinks_2025,xiao2024efficient,yu2024unveiling}. Similar behaviors have been observed in vision-language models, where background or visually salient patches dominate attention despite contributing little to the final prediction~\cite{woo2025don,kang2026see}. Together, these analyses suggest that attention is influenced not only by the input but also by systematic biases. In KB-VQA, this can lead models to focus on salient yet irrelevant tokens or regions instead of query-relevant evidence.

\begin{figure*}[t]
    \centering
    \includegraphics[width=\linewidth]{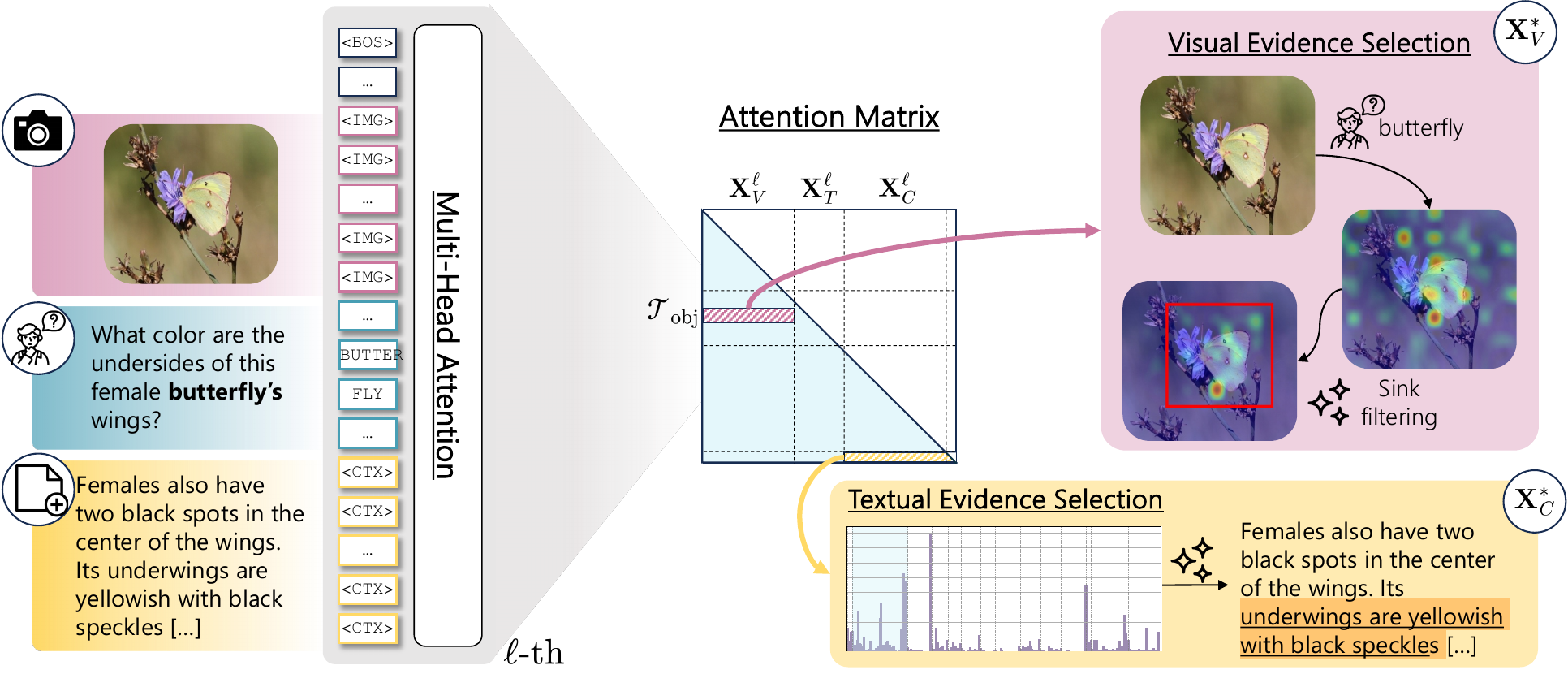}
    \vspace{-0.5cm}
    \caption{Overview of the proposed \oursl (\ours). \ours inspects the MLLM internal attention to identify relevant visual regions and textual evidence, discards irrelevant content, and highlights the selected cues for final answer generation.
    }
    \vspace{-0.3cm}
    \label{fig:model}
\end{figure*}

\vspace{-0.04cm}
\tit{Inference‑Level Attention Analysis}
A growing line of work modifies attention maps at inference time to improve grounding and reduce hallucinations in large models. In LLMs, attention scores identify relevant textual evidence~\cite{liu2025selfelicit}. In multimodal settings, some approaches identify dominant visual tokens and redistribute attention toward informative regions~\cite{kang2026see}. Others leverage internal attention dynamics to enhance semantically central visual regions during generation~\cite{lyu2026revealing_enhancing}, suppress disruptive instruction tokens that distort cross-modal alignment~\cite{chen2025attention_hijackers}, or reinforce high-focus attention patterns across heads to strengthen visual conditioning~\cite{zhang2024seeing_clearly}. 
Other approaches improve fine-grained visual perception by reprocessing localized image regions, deriving crops from the MLLM internal attention or gradient signals~\cite{ICLR2025_aaa0ac42}, constructing object-relevance maps from cached token representations~\cite{zhong2026focus}, or~performing confidence-guided hierarchical exploration~\cite{shen2025zoomeye}. 

In contrast, \ours{} jointly refines visual and retrieved textual evidence for knowledge-intensive VQA. It uses internal attention as a relevance signal to highlight query-relevant content and suppress distractions before generation, without modifying model parameters or architecture.

\section{Proposed Method}
\label{sec:method}

We introduce \oursl (\ours), a \textit{training-free} inference-time framework that improves multimodal evidence selection in MLLMs by \textit{highlighting} query-relevant cues in both the retrieved context and the input image. Our key idea is to treat the model internal attention distributions as an implicit relevance signal. Specifically, we (i) estimate which context tokens and visual regions the model considers relevant to the query, (ii) filter sink-driven attention patterns in the visual modality to improve evidence localization, (iii) suppress irrelevant or distracting content, and (iv) highlight the resulting textual evidence and image region through lightweight prompt-level markers during final answer generation. This approach requires no parameter updates and no architectural modifications. An overview of \ours can be seen in Fig.~\ref{fig:model}.

\subsection{Preliminaries}
\label{sec:preliminaries}

\tinytit{Task Definition}
We consider a pretrained MLLM that processes an image together with a textual prompt and generates answers autoregressively. Given an input image $I$ and a textual question $\mathbf{X}_T = [x_1,\dots,x_{N_T}]$, the model first converts the input image into a sequence of $N_V$ visual tokens, denoted as $\mathbf{X}_V
$, using its vision encoder and multimodal projector. 
In the KB-VQA setting, the model additionally receives retrieved textual evidence. A retrieval module selects the top-$n$ textual documents from a knowledge base (\eg, composed of Wikipedia articles), which are concatenated into a context sequence
$\mathbf{X}_C = [c_1,\dots,c_{N_C}]$.

The model processes a single multimodal sequence obtained by concatenating visual tokens, question tokens, and retrieved context, $\mathbf{X} = [\mathbf{X}_V;\mathbf{X}_T;\mathbf{X}_C]$ with total length $S = N_V + N_T + N_C$. During inference, the model generates an answer autoregressively conditioned on this sequence.

\tit{Attention Notation} 
To integrate information across the question, visual inputs, and retrieved knowledge, MLLMs rely on stacked self-attention layers that enable interactions between all tokens in the multimodal sequence. 

The input sequence is processed by a Transformer decoder with $L$ layers and $K$ attention heads. At layer $\ell$, the hidden states are denoted as $\mathbf{H}^{\ell} = [\mathbf{h}^{\ell}_1, \mathbf{h}^{\ell}_2, \dots, \mathbf{h}^{\ell}_S] \in \mathbb{R}^{S \times d}$, where $\mathbf{h}^{\ell}_i \in \mathbb{R}^{d}$ denotes the representation of token $i$ at layer $\ell$, and $d$ is the hidden dimension. The initial representation $\mathbf{H}^0$ corresponds to the embeddings of the concatenated multimodal input sequence $\mathbf{X}$.
Each layer applies causal multi-head self-attention~\cite{vaswani2017attention} to propagate information across tokens. We denote by $\mathbf{A}^{\ell,k} \in \mathbb{R}^{S \times S}$ the attention matrix at layer $\ell$ and head $k \in \{1,\dots,K\}$, where $\mathbf{A}^{\ell,k}[i,j]$ represents how strongly token $i$ attends to token $j$. Our method leverages attention patterns between tokens corresponding to the question, visual inputs, and retrieved context to estimate multimodal evidence relevance with respect to the query.

\subsection{Training-Free Multimodal Evidence Highlighting}
During an initial analysis stage, \ours generates a single token and uses the internal attention dynamics of the model to identify query-relevant visual regions and textual evidence. These cues are then explicitly exposed to the model during full answer generation. Our approach operates by extracting attention signals from the pretrained MLLM, filtering spurious activations caused by attention sinks while discarding irrelevant information, and highlighting the resulting evidence through lightweight prompt markers at inference time.

\tit{Self-Guided Visual Evidence Selection}
On the visual side, our objective is to identify the image regions that are truly relevant for answering the query by exploiting the model internal attention between the question tokens  $\mathbf{X}_T$ and the visual tokens $\mathbf{X}_V$ as an \textit{implicit relevance signal}.

Given the attention matrix $\mathbf{A}^{\ell,k} \in \mathbb{R}^{S\times S}$, we focus on the interactions between the target object in the question and the visual tokens. To this end, we use an off-the-shelf NLP library to identify the target-object tokens and extract the corresponding \textit{object-to-visual} attention submatrix:
\begin{equation}
\mathbf{A}^{\ell,k}_{\text{obj}\rightarrow\text{vis}} = \mathbf{A}^{\ell,k}[\mathcal{T}_{\text{obj}}, \mathcal{V}] \in \mathbb{R}^{|\mathcal{T}_{\text{obj}}|\times N_V},
\end{equation}
where $\mathcal{V} = \{1,\dots,N_V\}$ denotes the indices of visual tokens in the concatenated sequence and $\mathcal{T}_{\text{obj}} \subseteq \{N_V+1,\dots,N_V+N_T\}$ the indices of textual tokens referring to the target object. For example, given the question ``\textit{What color are the undersides of this female butterfly’s
wings?}'', $\mathcal{T}_{\text{obj}}$  corresponds to the token indices in the attention map associated with ``\textit{female butterfly}''. By restricting attention to these tokens, we explicitly capture how the object mentioned in the question interacts with the visual input, providing a focused measure of object-specific visual grounding.

Since cross-modal interactions are not uniformly localized but emerge across multiple layers and heads~\cite{shi2025vision, yin2025lifting}, and the target object may span multiple textual tokens, we aggregate object-to-visual attention signals across the object tokens, a subset of layers $L_{\text{vis}}$, and all attention heads. The resulting vector $\mathbf{a}_{\text{vis}}$ provides a single relevance score per visual token and serves as the initial spatial relevance signal:
\begin{equation}
\label{eq:a_vis}
    \mathbf{a}_{\text{vis}} =
    \frac{1}{|\mathcal{T}_{\text{obj}}| \cdot |L_\text{vis}| \cdot K  }
    \sum_{i \in \mathcal{T}_{\text{obj}}}
    \sum_{\ell \in L_\text{vis}}
    \sum_{k=1}^{K}
    {\mathbf{A}}_{\text{obj}\rightarrow\text{vis}}^{\ell,k} [i]
    \in \mathbb{R}^{N_V}.
\end{equation}

\begin{figure}[t]
    \centering
    \includegraphics[width=\linewidth]{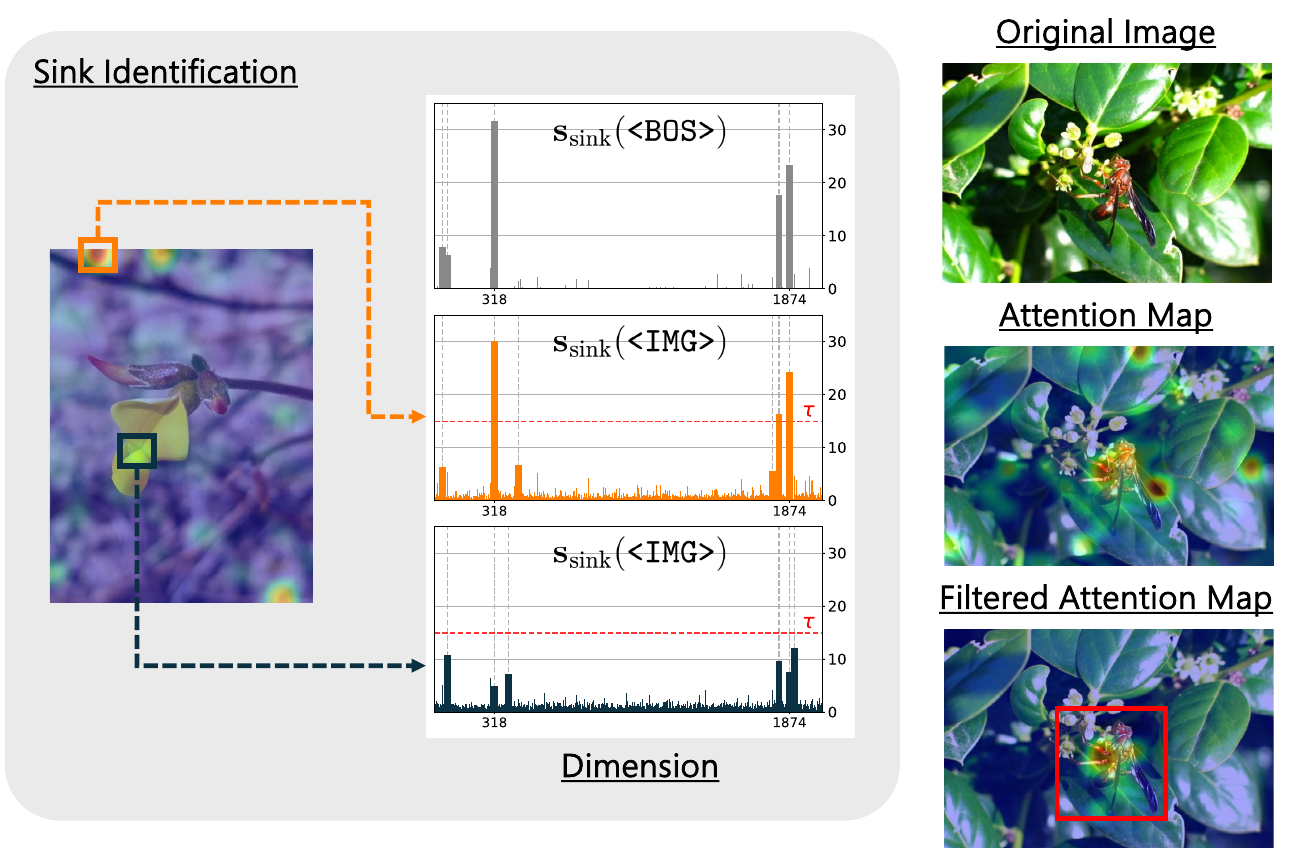}
    \vspace{-0.6cm}
    \caption{Qualitative examples of our visual evidence selection pipeline (right). In the attention map (left), irrelevant visual tokens (orange) exhibit disproportionately high activations in specific hidden-state dimensions, while relevant tokens (blue) remain moderate. The \texttt{BOS} token, known to act as an attention sink in LLMs, shows a similar pattern. In \ours, tokens with sink score exceeding a threshold $\tau$ are filtered.}
    \vspace{-0.3cm}
    \label{fig:visual_claim}
\end{figure}

\tit{Multi-Layer Attention Sink Filtering}
While $\mathbf{a}_{\text{vis}}$ captures object-conditioned visual relevance, Transformer models are known to exhibit \textit{attention sinks}~\cite{attention_sinks_2025,xiao2024efficient}: tokens that attract disproportionate attention mass regardless of their semantic contribution. This can produce spurious attention concentration and weaken grounding.
In multimodal settings, certain visual tokens similarly receive high attention across layers despite being spatially or semantically uninformative~\cite{kang2026see}, as shown in Fig.~\ref{fig:visual_claim} (right, middle row).

To distinguish sink-driven responses from genuine object relevance, we identify the hidden dimensions most strongly associated with attention sinks. Inspired by~\citet{kang2026see}, 
we analyze the hidden representations of visual tokens across a selected subset of layers $L_{\text{sink}}$.
For each layer $\ell \in L_{\text{sink}}$, let $\mathbf{H}^\ell \in \mathbb{R}^{S\times d}$ denote the hidden states and $\mathbf{H}^\ell_{\mathcal{V}} \in \mathbb{R}^{N_V\times d}$ the slice corresponding to visual tokens. We identify the hidden dimensions that consistently exhibit disproportionately high normalized activations for the \texttt{BOS} token in the base LLM, which is known to act as a prototypical attention sink~\cite{xiao2024efficient}.
The resulting set of sink dimensions, denoted as $\mathcal{D}_{\text{sink}} \subseteq \{1,\dots,d\}$, corresponds to a fixed subset of hidden dimensions determined by the LLM backbone. While the specific dimensions depend on the model architecture, they remain constant across input instances (Fig.~\ref{fig:visual_claim}, left). 

We define the attention sink score vector $\mathbf{s}_{\text{sink}}$ as the maximum absolute activation across $\mathcal{D}_{\text{sink}}$, normalized along the hidden dimension $d$, and averaged across the layers in $L_{\text{sink}}$:
\begin{equation}
\label{eq:sink_score}
\mathbf{s}_{\text{sink}} =
\frac{1}{|L_{\text{sink}}|}
\sum_{\ell \in L_{\text{sink}}}
\frac{
\max_{m \in \mathcal{D}_{\text{sink}}}
\left| \mathbf{H}^{\ell}_{\mathcal{V}}[:, m] \right|
}{
\|\mathbf{H}^{\ell}_{\mathcal{V}}\|_{\text{row}}
}
\in \mathbb{R}^{N_V},
\end{equation}
where $\|\mathbf{H}^{\ell}_{\mathcal{V}}\|_{\text{row}} = \sqrt{\frac{1}{d}\sum_\text{m=1}^{d}\mathbf{H}^\ell_\mathcal{V}[:, m]^2}$ is the row-wise RMS norm of $\mathbf{H}^{\ell}_{\mathcal{V}}$. Intuitively, $\mathbf{s}_{\text{sink}}$ quantifies, for each visual token, the activation along the sink dimensions relative to the overall magnitude of its hidden representation. Tokens with unusually high values in $\mathbf{s}_{\text{sink}}$ are therefore likely to be attention~sinks.

To suppress these artifacts, we filter visual tokens whose sink score exceeds a threshold $\tau$. Specifically, tokens where $\mathbf{s}_{\text{sink}} > \tau$ are marked as sinks and their corresponding entries in $\mathbf{a}_{\text{vis}}$ are set to zero. The resulting filtered visual relevance vector is reshaped to match the dimensions of the visual feature map, producing a 2D attention map $\mathbf{M}_{\text{vis}} \in \mathbb{R}^{H \times W}$ that reflects the model implicit grounding of the target object over image regions. Importantly, this filtering is applied only during the analysis of the first generated token; the attention maps used in the subsequent full generation step remain unchanged. Qualitative examples of the obtained attention maps are shown in Fig.~\ref{fig:visual_claim} (right, last row).

\tit{Bounding Box Extraction}
We now use the filtered attention map to localize the query entity and extract the corresponding image crop.
To convert the attention map into an explicit spatial region, we interpret $\mathbf{M}_{\text{vis}}$ as a spatial probability distribution by normalizing it. 
We then compute the weighted centroid $(c_x, c_y)$ of the attention map to locate the center of the highlighted visual region:
\begin{equation}
\label{eq:std_dev}
c_x = \sum_{h,w} w\,{\mathbf{M}}_{\text{vis}}(h,w), \quad
c_y = \sum_{h,w} h\,{\mathbf{M}}_{\text{vis}}(h,w).
\end{equation}
To quantify the spatial spread of the evidence, we compute the weighted standard deviations along each axis:
\begin{equation}
\label{eq:deviation}
\begin{aligned}
\sigma_x &= \sqrt{\sum_{h,w} (w-c_x)^2 {\mathbf{M}}_{\text{vis}}(h,w)}, \\
\sigma_y &= \sqrt{\sum_{h,w} (h-c_y)^2 {\mathbf{M}}_{\text{vis}}(h,w)}.
\end{aligned}
\end{equation}

The bounding box is finally defined as
\begin{equation}
\begin{aligned}
(x_1,y_1,x_2,y_2) =
(&c_x-\beta\sigma_x,\,
  c_y-\beta\sigma_y, \\
 &c_x+\beta\sigma_x,\,
  c_y+\beta\sigma_y),
\end{aligned}
\label{eq:centroid}
\end{equation}
which captures the region containing the most relevant visual evidence. The parameter $\beta$ controls the spatial extent of the bounding box, acting as a scaling factor that adjusts how much surrounding visual context is included.

\tit{Self-Guided Textual Evidence Selection}
Complementary to visual evidence localization, \ours extends the same self-guided selection strategy to the retrieved context. In fact, only a subset of the retrieved context $\mathbf{X}_C$ is relevant to the query~\cite{cocchi2025augmenting,yan2024echosight,yang2025omgm}, yet standard MLLMs often attend broadly across all context tokens, introducing noise and distracting reasoning~\cite{caffagni2024wiki}. Building on this observation, \ours leverages the model internal self-attention as an implicit relevance signal to distinguish informative from distracting sentences in $\mathbf{X}_C$. The resulting relevance estimates are then used to highlight the most informative evidence while discarding sentences deemed irrelevant before answer generation.

Intuitively, we measure how strongly the model focuses on each token in the retrieved context when generating the answer.
Specifically, we extract the \textit{last-to-context} attention submatrix,
corresponding to the attention from the last input token at position $t$ to all context tokens:
\begin{equation}
\mathbf{A}^{\ell,k}_{\text{last}\rightarrow\text{ctx}} =
\mathbf{A}^{\ell,k}[t, \mathcal{C}]
\in \mathbb{R}^{1 \times N_C},
\end{equation}
where $\mathcal{C} \subseteq \{N_V+N_T+1, \dots, N_V+N_T+N_C\}$ denotes the indices of the tokens corresponding to the retrieved context in the concatenated multimodal sequence.

As in the visual modality, to obtain a robust textual relevance score $\mathbf{a}_{\text{txt}}$, we aggregate these attention signals across multiple layers $L_\text{txt}$ and heads $K$, capturing diverse reasoning patterns learned at different depths of the model:
\begin{equation}
\label{eq:a_txt}
    \mathbf{a}_{\text{txt}} =
    \frac{1}{|L_\text{txt}| \cdot K}
    \sum_{\ell \in L_\text{txt}} \sum_{k=1}^{K}
    \mathbf{A}^{\ell,k}_{\text{last}\rightarrow\text{ctx}}
    \in \mathbb{R}^{N_C}.
\end{equation}

The token-level scores are then averaged within each sentence to obtain $\mathbf{\hat{a}}_{\text{txt}}$.Sentences whose relevance score exceeds a threshold $\alpha$ are selected as relevant textual evidence and explicitly highlighted. Conversely, sentences whose score falls below a lower threshold $\alpha_{\text{drop}}$ are considered irrelevant and removed from the retrieved context before answer generation. Sentences with scores between the two thresholds are retained without highlighting to preserve supporting context.

\tit{Inference with Evidence Refinement}
The final step of \ours refines the multimodal input by highlighting the identified relevant textual and visual evidence while discarding content considered irrelevant. This guides the MLLM toward the most informative cues and reduces the influence of potentially distracting information during answer generation.
To perform evidence highlighting efficiently, we adopt a prompt-augmentation strategy employing prompt-level markers. For visual evidence, we crop the image using the predicted bounding box and provide only the resulting region to the model enclosed by the markers
\texttt{\small<START\_IMPORTANT\_IMG>} and
\texttt{\small<END\_IMPORTANT\_IMG>}.
For textual evidence, the selected spans in $\mathbf{X}_C$ are wrapped with the markers
\texttt{\small<START\_IMPORTANT\_TXT>} and
\texttt{\small<END\_IMPORTANT\_TXT>}. Sentences identified as irrelevant are removed from the retrieved context, producing the refined textual input $\mathbf{X}_C^{*}$.
We update the task instructions to explicitly inform the model that the markers denote key evidence and should not appear in the output\footnote{We refer the reader to the supplementary material for the complete prompt templates.}. The model then generates the final response from the original question $\mathbf{X}_T$ and the refined multimodal input $\mathbf{X}_V^{*}, \mathbf{X}_C^{*}$, using the updated prompt template.

Notably, this inference-time mechanism leaves the model parameters unchanged and requires only minimal prompt modifications.
The complete method is detailed in Algorithm~1 reported in the supplementary material.

\section{Experiments}
\label{sec:experiments}

\subsection{Datasets}
We evaluate \ours on four KB-VQA benchmarks: Encyclopedic-VQA (E-VQA)~\cite{mensink2023encyclopedic}, InfoSeek~\cite{chen2023can}, ViQuAE~\cite{lerner2022viquae}, and OVEN~\cite{hu2023open}. E-VQA contains 221k question-answer pairs over 16.7k entities, including both single-hop and multi-hop questions, and is supported by a knowledge base of roughly 2M Wikipedia pages. InfoSeek comprises approximately 1.3M image-question-answer triplets spanning 11k entities, including previously unseen ones; its official setup uses 100k pages from a 6M-page Wikipedia knowledge base. We additionally report results on the ViQuAE test set and the OVEN validation set. Following prior work~\cite{cocchi2025augmenting,compagnoni2025reag,wenyi2025towards}, retrieval for ViQuAE and OVEN uses the E-VQA and InfoSeek knowledge bases, respectively. Together, these benchmarks span diverse scales, entity distributions, and reasoning challenges.

\setlength{\textfloatsep}{\origtextfloatsep}

\subsection{Implementation Details}

\tinytit{Models and Baselines}
We evaluate our method on a diverse set of recent MLLMs with different capacities. Specifically, we consider Qwen2-VL~\cite{wang2024qwen2} (2B, 7B), Qwen2.5-VL~\cite{bai2025qwen25vltechnicalreport} (3B, 7B, 32B), Qwen3-VL~\cite{bai2025qwen3} (4B, 8B), and InternVL3.5~\cite{wang2025internvl3} (4B, 8B, 38B). These models span multiple architectures and parameter scales, providing a broad set of models.

\begin{table}[t]
    \centering
\small
\setlength{\tabcolsep}{.18em}
\resizebox{\linewidth}{!}{
\begin{tabular}{l cc c ccc c c c c ccc}
\toprule
 & \multicolumn{2}{c}{\textbf{E-VQA}} & & \multicolumn{3}{c}{\textbf{InfoSeek}} & & \textbf{OVEN} && \multicolumn{1}{c}{\textbf{ViQuAE}}  \\
\cmidrule{2-3} \cmidrule{5-7} \cmidrule{9-9}  \cmidrule{11-11}
 & Single & All & & U-Q & U-E & All & & All && All & & \textbf{Avg} \\
\midrule
\rowcolor{TitleColor}
\multicolumn{14}{l}{\textit{Small-scale MLLMs (2-4B)}} \\
 $\blacktriangledown$ Qwen2-VL-2B & 17.5 & 16.0 & & 5.3 & 5.6 & 5.4 & & 1.2 && 15.0 & & 9.4 \\
\rowcolor{OurColor}
\hspace{0.4cm}\textbf{+ \ours (Ours)} & \textbf{20.5} & \textbf{18.5 }& & \textbf{21.8} & \textbf{21.4} & \textbf{21.6} & & \textbf{14.5} && \textbf{29.1} & & \textbf{20.9} & \inc{11.5} \\
\midrule
$\blacktriangledown$ Qwen2.5-VL-3B & 30.3 & 28.0 & & 22.6 & 22.2 & 22.4 & & 11.6 && 22.9 & & 21.2 \\
\rowcolor{OurColor}
\hspace{0.4cm}\textbf{+ \ours (Ours)} & \textbf{32.6} & \textbf{30.2} & & \textbf{30.3} & \textbf{29.8} & \textbf{30.1} & & \textbf{29.5} && \textbf{34.7} & & \textbf{31.1} & \inc{9.9} \\
\midrule
$\blacktriangledown$ Qwen3-VL-4B & 35.0 & 32.7 & & 28.0 & 28.6 & 28.3 & & 23.5 && 34.7 & & 29.8 \\
\rowcolor{OurColor}
\hspace{0.4cm}\textbf{+ \ours (Ours)} & \textbf{36.9} & \textbf{35.1} & & \textbf{29.9} & \textbf{30.3} & \textbf{30.1} & & \textbf{25.7} && \textbf{40.5} & & \textbf{32.9} & \inc{3.1} \\
\midrule
$\blacktriangledown$ InternVL3.5-4B  & 29.4 & 26.4 & & 28.8 & 29.1 & 29.0 & & 7.9 && 36.4 & & 24.9 \\
\rowcolor{OurColor}
\hspace{0.4cm}\textbf{+ \ours (Ours)} & \textbf{33.7} & \textbf{31.0} & & \textbf{29.9} & \textbf{29.0} & \textbf{29.5} & & \textbf{16.3} && \textbf{41.9} & & \textbf{29.7} & \inc{4.8} \\
\midrule
\rowcolor{TitleColor}
\multicolumn{14}{l}{\textit{Medium-scale MLLMs (7-8B)}} \\
$\blacktriangledown$ Qwen2-VL-7B & 25.6 & 22.9 & & 24.2 & 24.7 & 24.4 & & 11.1 && 33.0 & & 22.9  \\
\rowcolor{OurColor}
\hspace{0.4cm}\textbf{+ \ours (Ours)} & \textbf{29.6} & \textbf{26.6} & & \textbf{33.9} & \textbf{33.2} & \textbf{33.6} & & \textbf{30.5} && \textbf{50.6} & & \textbf{35.3} & \inc{12.5} \\
\midrule
$\blacktriangledown$ Qwen2.5-VL-7B & 32.1 & 30.2 & & 23.9 & 25.1 & 24.5 & & 20.2 && 36.4 & & 27.8 \\
\rowcolor{OurColor}
\hspace{0.4cm}\textbf{+ \ours (Ours)} & \textbf{35.0} & \textbf{32.2} & & \textbf{29.3} & \textbf{30.9} & \textbf{30.1} & & \textbf{30.0} && \textbf{50.5} & & \textbf{35.7} & \inc{7.9} \\
\midrule
$\blacktriangledown$ Qwen3-VL-8B  & 36.5 & 34.8 & & 29.1 & 30.4 & 29.7 & & 18.1 && 43.7 & & 31.6 \\
\rowcolor{OurColor}
\hspace{0.4cm}\textbf{+ \ours (Ours)} & \textbf{38.0} & \textbf{36.3} & & \textbf{32.7} & \textbf{32.2} & \textbf{32.5} & & \textbf{19.6} && \textbf{54.0} & & \textbf{35.6} & \inc{4.0} \\
\midrule
$\blacktriangledown$ InternVL3.5-8B  & 31.3 & 28.8 & & 29.4 & 29.8 & 29.6 & & 20.8 && 44.5 & & 30.9 \\
\rowcolor{OurColor}
\hspace{0.4cm}\textbf{+ \ours (Ours)} & \textbf{33.7} & \textbf{31.4} & & \textbf{31.6} & \textbf{31.1} & \textbf{31.3} & & \textbf{26.4} && \textbf{55.4} & & \textbf{36.1} & \inc{5.2} \\
\midrule
\rowcolor{TitleColor}
\multicolumn{14}{l}{\textit{Large-scale MLLMs (32-38B)}} \\
$\blacktriangledown$ Qwen2.5-VL-32B & 35.3 & 33.7 & & 26.7 & 26.1 & 26.4 & & 13.3 && 37.9 & & 27.8 \\
\rowcolor{OurColor}
\hspace{0.4cm}\textbf{+ \ours (Ours)} & \textbf{37.8} & \textbf{35.5} & & \textbf{29.6} & \textbf{28.5} & \textbf{29.0} & & \textbf{15.6} && \textbf{49.0} & & \textbf{32.3} & \inc{4.5} \\
\midrule
$\blacktriangledown$ InternVL3.5-38B  & 33.8 & 31.6 & & 33.1 & 33.0 & 33.1 & & 20.2 && 51.5 & & 34.1 \\
\rowcolor{OurColor}
\hspace{0.4cm}\textbf{+ \ours (Ours)} & \textbf{35.0} & \textbf{33.4} & & \textbf{33.6} & \textbf{33.2} & \textbf{33.4} & & \textbf{24.8} && \textbf{57.6} & & \textbf{37.3} & \inc{3.2} \\
\bottomrule
\end{tabular}
}
\vspace{-0.15cm}
\caption{Performance comparison on KB-VQA benchmarks. For each backbone, we compare the original zero-shot retrieval-augmented model with the same model augmented with \ours. Results are reported on the E-VQA test set, InfoSeek validation set, OVEN validation set, and ViQuAE test set. Avg denotes the average score across the four ``All'' columns; improvements over baseline are in green. 
}
\vspace{-0.4cm}
\label{tab:main_table}
\end{table}

\tit{Retrieval Pipeline}
We adopt a cross-modal entity retrieval pipeline to retrieve relevant knowledge for each input image. Following~\citet{yang2025omgm}, each Wikipedia entity in the knowledge base is represented by a concise textual summary. We encode all entity summaries using the EVA-CLIP text encoder~\cite{sun2024eva}. Given an input image, we obtain its representation with the EVA-CLIP visual encoder and perform image-to-text retrieval through inner-product similarity search with FAISS~\cite{johnson2019billion}. The top-$n$ retrieved documents are concatenated to form the textual context provided to the MLLM, with $n=3$ in our experiments. When re-ranking is enabled, we retrieve the top-$3$ documents and retain the top-$5$ sections for answer generation.

\tit{Evidence Selection}
During generation of the first output token, we analyze attention patterns across different decoder layer ranges, selecting modality-specific subsets corresponding to where MLLMs typically process each modality. Prior work indicates that textual evidence is primarily consolidated in deeper layers, whereas visual grounding emerges in intermediate layers~\cite{kang2025your,liu2025selfelicit,jiang2025devils}. Attention sinks, instead, may arise throughout the network. 
Accordingly, we estimate textual relevance from the second half of the decoder layers, denoted by $L_{\text{txt}}$. Following~\citet{compagnoni2025reag}, we use spaCy for lightweight target-object extraction. Visual relevance is computed over the middle half of the decoder, spanning one-quarter to three-quarters of the total depth and denoted by $L_{\text{vis}}$. We perform sink detection over the same intermediate range, setting $L_{\text{sink}} = L_{\text{vis}}$. $D_\text{sink}$ is found following~\citet{kang2026see}.

During sink filtering, the threshold $\tau$ is set to the 25th percentile of the sink scores computed over all visual tokens.  
For textual evidence refinement, we highlight sentences whose attention score exceeds half of the maximum sentence-level score, setting $\alpha = 0.5 \cdot \max(\mathbf{\hat{a}}_{\text{txt}})$.
We discard sentences whose score is lower than one third of the highlighting threshold, using $\alpha_{\text{drop}} = \alpha/3$.
Sentences whose scores lie between $\alpha_{\text{drop}}$ and $\alpha$ are retained without highlighting. For InternVL3.5 backbones, we use a more conservative filtering criterion and halve the default dropping threshold. Finally, in the bounding box extraction step (cf. Eq.~\ref{eq:centroid}) we set $\beta = 2$. Further details on design choices, sink dimension and selection of hyperparameter values are provided in the supplementary~material.

\begin{figure}[t]
    \centering
    \includegraphics[width=\linewidth]{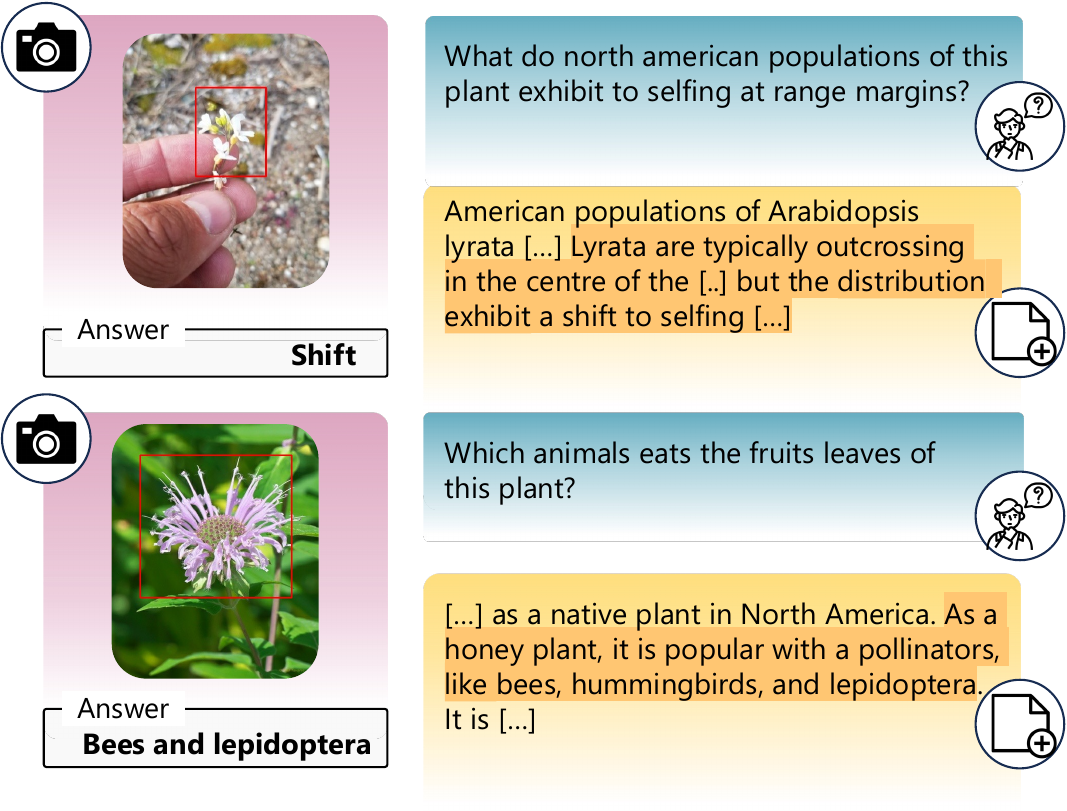}
    \vspace{-0.6cm}
    \caption{Qualitative examples of \ours highlighting query-relevant visual regions and textual evidence, enabling the model to generate the correct answer.}
    \vspace{-0.35cm}
    \label{fig:qual_long}
\end{figure}

\subsection{Experimental Results}

\tinytit{Main Results}
Table~\ref{tab:main_table} reports the performance of \ours on E-VQA, InfoSeek, OVEN, and ViQuAE. For each backbone, we compare the original zero-shot retrieval-augmented MLLM, where all retrieved passages are appended to the input context, with the same model augmented with \ours, which refines both visual and textual evidence before answer generation. 

Across all model families, scales, and benchmarks, \ours consistently improves performance. The largest average gains are observed for Qwen2-VL-7B and Qwen2-VL-2B, which improve by $+12.5$ and $+11.5$ points, respectively. \ours also yields substantial improvements for Qwen2.5-VL-3B ($+9.9$) and Qwen2.5-VL-7B ($+7.9$), demonstrating that the benefits of multimodal evidence refinement are not restricted to a specific architecture. 
The improvements are particularly pronounced on benchmarks requiring effective reasoning over retrieved knowledge.
For instance, Qwen2-VL-7B improves from $11.1$ to $30.5$ on OVEN and from $33.0$ to $50.6$ on ViQuAE. Gains also remain consistent for~larger models, with InternVL3.5-38B increasing from $34.1$ to $37.3$ on average.

\begin{table}[t]
    \centering
    \small
    \setlength{\tabcolsep}{.2em}
    \resizebox{\linewidth}{!}{%
    \begin{tabular}{l c c c c c cc}
        \toprule
        & \multicolumn{2}{c}{\textbf{E-VQA}}
        & & \multicolumn{2}{c}{\textbf{InfoSeek}}
        & \\
        \cmidrule{2-3}
        \cmidrule{5-6}
        & Base & +\textbf{\ours}
        & & Base & +\textbf{\ours}
        & \\
        \midrule

        \rowcolor{TitleColor}
        $\blacktriangledown$ Qwen2.5-VL-3B
        & 28.0 & \cellcolor{OurColor}\textbf{30.2}
        & & 22.4 & \cellcolor{OurColor}\textbf{30.1}
        & & \inc{4.95} \\

        \rowcolor{TitleColor}
        \hspace{0.3cm}{+ CoT~\cite{wei2022chain}}
        & 28.1 & \cellcolor{OurColor}\textbf{30.1}
        & & 29.5 & \cellcolor{OurColor}\textbf{30.1}
        & & \inc{1.30} \\

        \rowcolor{TitleColor}
        \hspace{0.3cm}{+ Critic~\cite{compagnoni2025reag}}
        & 30.1 & \cellcolor{OurColor}\textbf{31.3}
        & & 30.6 & \cellcolor{OurColor}\textbf{31.1}
        & & \inc{0.85} \\

        \hspace{0.3cm}{+ Critic~\cite{compagnoni2025reag}}
        & 33.3 & \cellcolor{OurColor}\textbf{34.4}
        & & 33.5 & \cellcolor{OurColor}\textbf{34.4}
        & & \inc{1.00} \\
        
        \hspace{0.3cm}{+ EchoSight Re-ranker~\cite{yan2024echosight}}
        & 29.8 & \cellcolor{OurColor}\textbf{31.1}
        & & 30.7 & \cellcolor{OurColor}\textbf{32.0}
        & & \inc{1.30} \\

        \hspace{0.3cm}{+ Qwen3-VL-Reranker-8B~\cite{li2026qwen3}}
        & 30.5 & \cellcolor{OurColor}\textbf{31.6}
        & & \textbf{33.6} & \cellcolor{OurColor}\textbf{33.6}
        & & \inc{0.55} \\

        \midrule

        \rowcolor{TitleColor}
        $\blacktriangledown$ Qwen2.5-VL-7B
        & 30.2 & \cellcolor{OurColor}\textbf{32.2}
        & & 24.5 & \cellcolor{OurColor}\textbf{30.1}
        & & \inc{3.80} \\

        \rowcolor{TitleColor}
        \hspace{0.3cm}{+ CoT~\cite{wei2022chain}}
        & 28.6 & \cellcolor{OurColor}\textbf{32.1}
        & & 28.5 & \cellcolor{OurColor}\textbf{30.1}
        & & \inc{2.55} \\

        \rowcolor{TitleColor}
        \hspace{0.3cm}{+ Critic~\cite{compagnoni2025reag}}
        & 30.1 & \cellcolor{OurColor}\textbf{30.6}
        & & 29.5 & \cellcolor{OurColor}\textbf{29.8}
        & & \inc{0.40} \\

        \hspace{0.3cm}{+ Critic~\cite{compagnoni2025reag}}
        & 35.5 & \cellcolor{OurColor}\textbf{36.3}
        & & 34.0 & \cellcolor{OurColor}\textbf{35.2}
        & & \inc{1.00} \\

        \hspace{0.3cm}{+ EchoSight Re-ranker~\cite{yan2024echosight}}
        & 32.9 & \cellcolor{OurColor}\textbf{34.3}
        & & 30.4 & \cellcolor{OurColor}\textbf{32.8}
        & & \inc{1.90} \\

        \hspace{0.3cm}{+ Qwen3-VL-Reranker-8B~\cite{li2026qwen3}}
        & 33.6 & \cellcolor{OurColor}\textbf{34.0}
        & & 32.6 & \cellcolor{OurColor}\textbf{34.0}
        & & \inc{0.90} \\
        
        \midrule

        \rowcolor{TitleColor}
        $\blacktriangledown$ Qwen2.5-VL-32B
        & 33.7 & \cellcolor{OurColor}\textbf{35.5}
        & & 26.4 & \cellcolor{OurColor}\textbf{29.0}
        & & \inc{2.20} \\

        \rowcolor{TitleColor}
        \hspace{0.3cm}{+ CoT~\cite{wei2022chain}}
        & 32.8 & \cellcolor{OurColor}\textbf{35.6}
        & & \textbf{29.7} & \cellcolor{OurColor}29.3
        & & \inc{1.20} \\

        \rowcolor{TitleColor}
        \hspace{0.3cm}{+ Critic~\cite{compagnoni2025reag}}
        & 35.7 & \cellcolor{OurColor}\textbf{37.0}
        & & 29.8 & \cellcolor{OurColor}\textbf{30.9}
        & & \inc{1.20} \\

        \hspace{0.3cm}{+ Critic~\cite{compagnoni2025reag}}
        & 37.9 & \cellcolor{OurColor}\textbf{40.7}
        & & 34.4 & \cellcolor{OurColor}\textbf{35.2}
        & & \inc{1.80} \\

        \hspace{0.3cm}{+ EchoSight Re-ranker~\cite{yan2024echosight}}
        & 36.9 & \cellcolor{OurColor}\textbf{38.4}
        & & 28.8 & \cellcolor{OurColor}\textbf{30.0}
        & & \inc{1.35} \\

        \hspace{0.3cm}{+ Qwen3-VL-Reranker-8B~\cite{li2026qwen3}}
        & 36.8 & \cellcolor{OurColor}\textbf{38.6}
        & & 31.0 & \cellcolor{OurColor}\textbf{31.3}
        & & \inc{1.05} \\

        \bottomrule
    \end{tabular}
    }
    \vspace{-0.15cm}
    \caption{Performance comparison of inference-time strategies combined with \ours. Gray rows denote zero-shot strategies that require no additional
    trained components.}
    \label{tab:training_free_combinations}
    \vspace{-0.3cm}
\end{table}

Overall, the average improvement ranges from $+3.1$ to $+12.5$ points across the evaluated models, showing that 
attention-guided evidence highlighting is a general and robust mechanism, yielding consistent and substantial accuracy gains regardless of backbone capacity.

Fig.~\ref{fig:qual_long} presents qualitative examples of \ours on E-VQA. As shown, the method reliably identifies the precise image region containing the object referenced in the question, while in the retrieved textual context, it highlights the sentence that directly provides the answer, demonstrating the effectiveness of \ours in highlighting the relevant multimodal evidence.

\tit{Integration with Existing Inference Strategies}
Table~\ref{tab:training_free_combinations} evaluates the compatibility of \ours with a diverse set of KB-VQA inference strategies. We consider training-free methods, including chain-of-thought prompting (CoT)~\cite{wei2022chain} and a prompt-based critic model following~\citet{compagnoni2025reag}, where the same zero-shot MLLM used for answer generation is prompted to retain or discard candidate passages. We additionally evaluate the task-specific critic model released by~\citet{compagnoni2025reag}, which was fine-tuned for passage filtering, as well as specialized re-ranking models~\cite{li2026qwen3,yan2024echosight}\footnote{All configurations use $n=3$ retrieved documents, except the trained Critic, which follows its original $n=20$ setting.}. Across these settings, \ours generally provides complementary gains, reaching up to $+2.55$ points when combined with existing inference strategies and up to $+4.95$ points for the standard retrieval-augmented baseline. These results show that multimodal evidence refinement remains beneficial even when reasoning, passage filtering, or re-ranking is already applied.

\tit{Computational Analysis}
\ours limits memory overhead by monkey-patching the model forward to compute only the~attention scores required for evidence selection, while retaining FlashAttention-2 for all remaining attention computations.
On E-VQA with~Qwen2.5-VL-3B, 
it reduces the final input by $71.93$\%: $-124.0$ visual tokens ($-54.24$\%) and $-2,139.3$ textual tokens ($-72.90$\%). 
Considering GPU-side model execution, \ours adds $14.1$\% latency overhead over the baseline and a $+42.8$\% in FLOPs.
However, the refined input reduces generation latency by 18.1\% and generation FLOPs by 81.1\%.

\subsection{Ablation Studies}

\tinytit{Effectiveness of Multimodal Highlighting}
Table~\ref{tab:ablation1} isolates the contribution of visual and textual highlighting on the 3B and 7B versions of Qwen2.5-VL. As shown, each modality improves over the corresponding baseline in most settings, showing that both visual localization and textual evidence selection contribute to the final performance. Combining both in \ours consistently yields the best results, reaching $32.2$ on E-VQA, $30.1$ on InfoSeek, $30.0$ on OVEN, and $50.5$ on ViQuAE for the 7B backbone. 
This confirms that visual and textual evidence refinement are complementary, with their joint application consistently yielding the strongest overall performance across the different evaluated settings.

\begin{table}[t]
    \centering
\small
\setlength{\tabcolsep}{.2em}
\resizebox{\linewidth}{!}{
\begin{tabular}{l cc c cc c ccc c c c c}
\toprule
& \multicolumn{2}{c}{\textbf{Highlight}} & & \multicolumn{2}{c}{\textbf{E-VQA}} & & \multicolumn{3}{c}{\textbf{InfoSeek}} & & \textbf{OVEN} && \multicolumn{1}{c}{\textbf{ViQuAE}}  \\
\cmidrule{2-3} \cmidrule{5-6} \cmidrule{8-10} \cmidrule{12-12} \cmidrule{14-14} 
 & Visual & Textual & & Single & All & & U-Q & U-E & All & & All && All \\
\midrule
\rowcolor{TitleColor}
$\blacktriangledown$ Qwen2.5-VL-3B & - & - & & 30.3 & 28.0 & & 22.6 & 22.2 & 22.4 & & 11.6 && 22.9 \\
& \cmark & - & & 31.6 & 29.4 & & 24.1 & 23.7 & 23.9 & & 16.6 & & 23.6 \\
& - & \cmark & & 31.7 & 29.1 & & {29.8} & {29.2} & {29.5} & & 28.0 & & 34.6 \\
\rowcolor{OurColor}
\hspace{0.3cm}\textbf{+ \ours (Ours)} & \cmark & \cmark & & \textbf{32.6} & \textbf{30.2} & & \textbf{30.3} & \textbf{29.8} & \textbf{30.1} & & \textbf{29.5} & & \textbf{34.7} \\
\midrule
\rowcolor{TitleColor}
$\blacktriangledown$ Qwen2.5-VL-7B & - & - & & 32.1 & 30.2 & & 23.9 & 25.1 & 24.5 & & 20.2 && 36.4 \\
& \cmark & - & & 32.4 & 30.2 & & 24.4 & 25.4 & 24.9 & & 21.0 & & 34.3 \\
& - & \cmark & & 34.7 & 32.1 & & 28.8 & 29.9 & 29.4 & & 28.7 & & 49.6 \\
\rowcolor{OurColor}
\hspace{0.3cm}\textbf{+ \ours (Ours)} & \cmark & \cmark & & \textbf{35.0} & \textbf{32.2} & & \textbf{29.3} & \textbf{30.9} & \textbf{30.1} & & \textbf{30.0} & & \textbf{50.5} \\
\bottomrule
\end{tabular}
}
\vspace{-0.15cm}
\caption{Ablation evaluating visual and textual highlighting independently and jointly across KB-VQA benchmarks.
}
\vspace{-0.4cm}
\label{tab:ablation1}
\end{table}

\begin{figure}[t]
    \centering
    \includegraphics[width=\linewidth]{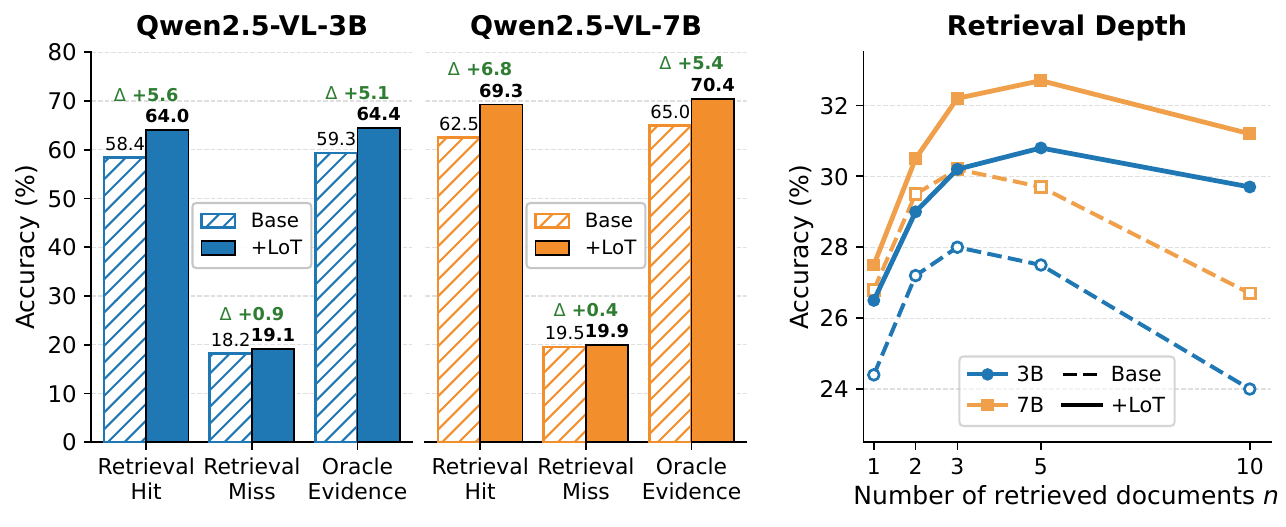}
    \vspace{-0.7cm}
    \caption{Performance on E-VQA~(All) under different evidence-availability (left) and retrieval-depth settings (right). 
    }
    \label{fig:plots}
    \vspace{-0.55cm}
\end{figure}

\tit{Evidence Availability Analysis}
To assess how \ours exploits retrieved evidence, we evaluate E-VQA separately when the retriever is able to find the answer-bearing document (\emph{retrieval hit}
) and when it does not (\emph{retrieval miss}).
As shown in Fig.~\ref{fig:plots} (left),  \ours achieves substantial gains of $+5.6$ and $+6.8$ on retrieval hits, bringing performance close to the oracle setting and demonstrating its ability to identify and emphasize the relevant evidence when available. On retrieval misses, gains are smaller but remain positive ($+0.9$ and $+0.4$): jointly highlighting the most informative visual cues and removing the least relevant sentences, \ours reduces noise and helps the model rely more effectively on the image.

\tit{Effectiveness with Oracle Evidence}
To isolate the effect of evidence utilization from retrieval quality, in Fig.~\ref{fig:plots} (left) we also evaluate the oracle setting on E-VQA in which each sample is provided only with its ground-truth Wikipedia page. \ours{} improves accuracy from $59.3$ to $64.4$ for the 3B model and from $65.0$ to $70.4$ for the 7B model. Notably, even when provided with only the correct page and minimal distracting context, \ours{} still delivers substantial gains.

\tit{Effect of Retrieval Depth}
In Fig.~\ref{fig:plots} (right), we vary the number of retrieved documents from $1$ to $10$ on E-VQA. \ours{} consistently outperforms the corresponding baselines. Notably, as $n$ increases from $3$ to $5$, \ours{} benefits from the higher likelihood of retrieving the answer-bearing evidence, whereas both baselines degrade under additional retrieval~noise.

\section{Conclusion}
\label{sec:conclusion}
We introduced \ours, a training-free inference-time framework that uses internal attention signals to refine visual and textual evidence before answer generation. Across four KB-VQA benchmarks and ten off-the-shelf MLLMs, \ours consistently improves zero-shot performance and complements existing refinement strategies, demonstrating that model-internal attention can enhance multimodal evidence utilization without additional training, auxiliary models, or architectural changes.

\section*{Acknowledgments}
We acknowledge the CINECA award under the ISCRA initiative, for the availability of high-performance computing resources. This work has been supported
by the EU Horizon project “ELLIOT - European Large Open Multi-Modal Foundation Models For Robust Generalization On Arbitrary Data Streams” (No.
101214398) and by the EuroHPC JU project “MINERVA” (GA No. 101182737).

\bigskip

\bibliography{bibliography}

\clearpage

\clearpage
\twocolumn[
\begin{center}
    {\LARGE\bfseries Supplementary Material}
\end{center}
\vspace{1.5em}
]

\section{\ours Algorithm}

Algorithm~\ref{alg:ret} summarizes the \ours pipeline. Given an image, a question, and retrieved context, \ours first generates a single output token to extract object-to-visual and last-to-context attention scores across decoder layers and heads. The aggregated visual attention is filtered using the detected sink dimensions and converted into a query-specific bounding box. For text, token-level attention is averaged within each sentence to estimate its relevance: highly relevant sentences are selected for highlighting, low-relevance sentences are discarded, and intermediate-scoring sentences are retained unchanged. The selected crop and textual evidence are then used to reconstruct a multimodal input with dedicated markers highlighting the most informative content. Finally, this refined input is used to generate the final answer.

\setlength{\textfloatsep}{7pt}
\begin{algorithm}[t]
\caption{\oursl (\ours): Training-Free Multimodal Evidence Highlighting}
\label{alg:ret}
\begin{algorithmic}[1]
\algrenewcommand\alglinenumber[1]{}
\State \textbf{Input:} image $\mathbf{X}_V$, question $\mathbf{X}_T$, retrieved context $\mathbf{X}_C$

\algrenewcommand\alglinenumber[1]{\scriptsize #1}

\State $\mathbf{A} \leftarrow$ MLLM$([\mathbf{X}_V;\mathbf{X}_T;\mathbf{X}_C])$ \hfill $\triangleright$ first forward pass to obtain attention matrix

\For{each $(\ell, k)$}

\State \hlpink{$\mathbf{A}^{\ell,k}_{\text{obj}\rightarrow\text{vis}}$} 
$\leftarrow \mathbf{A}^{\ell,k}[\mathcal{T}_{\text{obj}}, \mathcal{V}]$
\hfill $\triangleright$ object-to-visual attention

\State \hlyellow{$\mathbf{A}^{\ell,k}_{\text{last}\rightarrow\text{ctx}}$} 
$\leftarrow \mathbf{A}^{\ell,k}[t, \mathcal{C}]$
\hfill $\triangleright$ last-to-context attention

\EndFor

\State $\mathbf{a}_{\text{vis}} \leftarrow$ average over $k, \ell, \mathcal{T}_{\text{obj}}$
\hfill  $\triangleright$ Eq.~2

\State $\mathbf{a}_{\text{txt}} \leftarrow$ average over $k, \ell$
\hfill $\triangleright$ Eq.~8

\algrenewcommand\alglinenumber[1]{}
\State \hlpink{Visual Evidence Selection}
\algrenewcommand\alglinenumber[1]{\scriptsize #1}

\State $\mathbf{s}_\text{sink} \leftarrow$ compute sink score vector using $\mathcal{D}_{\text{sink}}$
\hfill $\triangleright$ Eq.~3

\State $\mathbf{a}_{\text{vis}} \leftarrow \mathbf{a}_{\text{vis}}$ where $\mathbf{s}_\text{sink} < \tau$ else 0 \hfill $\triangleright$ sink filtering

\State $\mathbf{M}_{\text{vis}} \leftarrow$ reshape $\mathbf{a}_{\text{vis}}$ in 2D attention map

\State $(c_x,c_y),(\sigma_x,\sigma_y) \leftarrow \mathbf{M}_{\text{vis}}$
\hfill $\triangleright$~centroid~and~spread (Eq.~4,~5)

\State $(x_1,y_1,x_2,y_2) \leftarrow
(c_x-\beta\sigma_x,\; c_y-\beta\sigma_y,\; c_x+\beta\sigma_x,\; c_y+\beta\sigma_y)$
\hfill $\triangleright$ evidence bbox

\algrenewcommand\alglinenumber[1]{}
\State \hlyellow{Textual Evidence Selection}
\algrenewcommand\alglinenumber[1]{\scriptsize #1}

\State $\mathbf{\hat{a}}_{\text{txt}} \leftarrow \text{avg}(\text{NLP}(\mathbf{a}_{\text{txt}}))$
\hfill $\triangleright$ average over sentence

\State  $\mathbf{X}_C \leftarrow \mathbf{X}_C[\mathbf{\hat{a}}_{\text{txt}} > \alpha_{drop}]$
\hfill $\triangleright$ discard irrelevant sentences

\State  
Identify sentences where $\mathbf{\hat{a}}_{\text{txt}} > \alpha$
\hfill $\triangleright$ textual evidence

\algrenewcommand\alglinenumber[1]{}
\State \hlours{Multimodal Evidence Highlighting}
\algrenewcommand\alglinenumber[1]{\scriptsize #1}

\State $\mathbf{X}_V^*, \mathbf{X}_C^* \leftarrow$ add evidence markers

\State answer $\leftarrow$ MLLM$([\mathbf{X}_V^*;\mathbf{X}_T;\mathbf{X}_C^*])$ \hfill $\triangleright$ generate answer on highlighted evidence
\end{algorithmic}
\end{algorithm}

\section{Additional Experimental Results}

\subsection{Additional Ablation Studies}

\begin{table}[t]
    \centering
\small
\setlength{\tabcolsep}{.28em}
\resizebox{0.95\linewidth}{!}{
\begin{tabular}{lc cc c ccc}
\toprule
& & \multicolumn{2}{c}{\textbf{E-VQA}}
& & \multicolumn{3}{c}{\textbf{InfoSeek}} \\
\cmidrule{3-4} \cmidrule{6-8}
& & Single-Hop & All
& & Unseen-Q & Unseen-E & All \\
\midrule

\rowcolor{TitleColor}
$\blacktriangledown$ Qwen2.5-VL-3B
& & 30.3 & 28.0
& & 22.6 & 22.2 & 22.4 \\
\hspace{0.4cm}Random
& & 22.2 & 21.2
& & 22.2 & 22.1 & 22.1 \\
\hspace{0.4cm}Leading
& & 24.1 & 23.0
& & 28.2 & 28.6 & 28.4 \\
\hspace{0.4cm}BM25
& & 30.6 & 28.2
& & 28.5 & 28.0 & 28.3 \\
\hspace{0.4cm}Retrieval Round-Robin
& & 22.6 & 21.7
& & 24.2 & 25.2 & 24.7 \\
\hspace{0.4cm}Prompted (Two-Pass)
& & 30.5 & 28.2
& & 28.7 & 28.2 & 28.4 \\
\rowcolor{OurColor}
\hspace{0.4cm}\textbf{\ours{} (Textual)}
& & \textbf{31.7} & \textbf{29.1}
& & \textbf{29.8} & \textbf{29.2} & \textbf{29.5} \\

\midrule

\rowcolor{TitleColor}
$\blacktriangledown$ Qwen2.5-VL-7B
& & 32.1 & 30.2
& & 23.9 & 25.1 & 24.5 \\
\hspace{0.4cm}Random
& & 23.9 & 23.3
& & 22.6 & 23.5 & 23.1 \\
\hspace{0.4cm}Leading
& & 26.9 & 25.8
& & 27.3 & 29.6 & 28.4 \\
\hspace{0.4cm}BM25
& & 34.5 & 31.9
& & 27.5 & 28.9 & 28.2 \\
\hspace{0.4cm}Retrieval Round-Robin
& & 25.5 & 24.4
& & 24.0 & 25.9 & 24.9 \\
\hspace{0.4cm}Prompted (Two-Pass)
& & 33.6 & 31.4
& & 27.4 & 28.8 & 28.1 \\
\rowcolor{OurColor}
\hspace{0.4cm}\textbf{\ours{} (Textual)}
& & \textbf{34.7} & \textbf{32.1}
& & \textbf{28.8} & \textbf{29.9} & \textbf{29.4} \\

\midrule
\midrule

\rowcolor{TitleColor}
$\blacktriangledown$ Qwen2.5-VL-3B
& & 30.3 & 28.0
& & 22.6 & 22.2 & 22.4 \\
\hspace{0.4cm}Center Crop
& & 31.2 & 29.2
& & 23.9 & 23.2 & 23.5 \\
\hspace{0.4cm}Random Crop
& & 31.1 & 29.1
& & 22.1 & 21.6 & 21.9 \\
\rowcolor{OurColor}
\hspace{0.4cm}\textbf{\ours{} (Visual)}
& & \textbf{31.6} & \textbf{29.4}
& & \textbf{24.1} & \textbf{23.7} & \textbf{23.9} \\

\midrule

\rowcolor{TitleColor}
$\blacktriangledown$ Qwen2.5-VL-7B
& & 32.1 & \textbf{30.2}
& & 23.9 & 25.1 & 24.5 \\
\hspace{0.4cm}Center Crop
& & 31.8 & 29.9
& & 23.3 & 24.3 & 23.8 \\
\hspace{0.4cm}Random Crop
& & 31.6 & 29.5
& & 23.3 & 24.3 & 23.8 \\
\rowcolor{OurColor}
\hspace{0.4cm}\textbf{\ours{} (Visual)}
& & \textbf{32.4} & \textbf{30.2}
& & \textbf{24.4} & \textbf{25.4} & \textbf{24.9} \\

\bottomrule
\end{tabular}
}
\vspace{-0.15cm}
\caption{Comparison of \ours with budget-matched non-attention-based selection strategies on E-VQA and InfoSeek. Textual methods retain $30$\% of the context, while visual methods retain $46$\% of the image area.}
\label{tab:causal_selector_baselines}
\end{table}

\tinytit{Comparison with Non-Attention-Based Selection Strategies}
To determine whether the gains of \ours arise simply from shortening the textual context or cropping the image,  Table~\ref{tab:causal_selector_baselines} compares \ours against several non-attention-based selection strategies operating under the same input budget. For all textual controls, the visual input is left unchanged and the selector targets a token budget corresponding to $30$\% of the valid retrieved context. \textit{Random} constructs a deterministic, sample-specific random ordering of the candidate sentences and retains sentences from this ordering until the token budget is reached. \textit{Leading} instead preserves the original context order and retains the earliest sentences that fit within the budget. \textit{BM25} ranks sentences by their lexical relevance to the question, using term-frequency saturation, inverse document frequency, and length normalization. The BM25 statistics are computed over the sentences in the retrieved context of each example, and the highest-ranked sentences are retained. \textit{Retrieval Round-Robin} follows the original ranking of the retrieved passages and interleaves their sentences: it considers the first sentence from each passage in retrieval order, followed by the second sentence from each passage, and so forth, until the token budget is reached. Finally, \textit{Prompted (Two-Pass)} asks the same MLLM, conditioned on the image and question, to return a ranked list of the sentence IDs that are most useful for answering the question. The model then performs a second generation pass using only the selected textual evidence and the original image. For vision, \textit{Center Crop} selects a central region covering $46$\% of the image area, whereas \textit{Random Crop} selects an equally sized region at a random location. These visual controls operate on the original textual context.

For text, random and retrieval-based selectors often substantially reduce performance, while stronger lexical and prompted selectors yield more competitive results. Nevertheless, \ours achieves the best overall accuracy on both datasets and model sizes. This indicates that its improvements can not be explained solely by retaining fewer tokens, but depend on selecting evidence according to the query-conditioned relevance signals of the model.
A similar pattern emerges for visual selection. Generic center and random crops provide inconsistent benefits and can reduce performance, particularly for the 7B model. Overall, these comparisons show that targeted multimodal evidence selection is more effective than generic input compression at an equivalent evidence budget.

\tit{Comparison with Different Visual Highlighting Strategies}
Table~\ref{tab:ablation2} compares alternative strategies for presenting the
selected visual evidence. Beyond the baseline, which uses only the original
input image, we consider drawing the estimated bounding box on the original
image (\textit{w/ bbox on input image}), providing only the selected crop
(\textit{w/ only selected evidence}), and using both the original image
and the crop either without markers
(\textit{w/ input image and evidence (no highlight)}) or with highlighting
markers around both images
(\textit{w/ highlight on both input image and evidence}). The complete
\ours strategy instead provides only the selected crop enclosed by
highlighting markers (\textit{highlight on evidence only}).

Drawing the bounding box on the original image provides no consistent benefit,
while using only the selected crop yields at most limited improvements.
Providing both the original image and the crop without markers generally
degrades performance, suggesting that duplicated visual content introduces
additional noise. Highlighting both images is more effective, but remains
inconsistent across models and datasets. Overall, \ours achieves the best results in all settings, indicating that explicitly highlighting
the localized evidence is more effective than preserving or duplicating the
full image. This design also avoids the additional visual tokens required to
process the original image.

\begin{table}[t]
    \centering
\small
\setlength{\tabcolsep}{.15em}
\resizebox{\linewidth}{!}{
\begin{tabular}{lc cc c ccc c}
\toprule
& & \multicolumn{2}{c}{\textbf{E-VQA}} & & \multicolumn{3}{c}{\textbf{InfoSeek}} \\
\cmidrule{3-4} \cmidrule{6-8}
 & & Single & All & & U-Q & U-E & All \\
\midrule
\rowcolor{TitleColor}
$\blacktriangledown$ Qwen2.5-VL-3B (only input image) & & 30.3 & 28.0 & & 22.6 & 22.2 & 22.4 \\
\hspace{0.4cm}w/ bbox on input image & & 29.6 & 27.4 & & 22.8 & 22.2 & 22.5 \\
\hspace{0.4cm}w/ only selected evidence & & 30.7 & 28.3 & & 22.7 & 22.3 & 22.5 \\
\hspace{0.4cm}w/ input image and evidence (no highlight) & & 30.0 & 27.8 & & 21.4 & 20.8 & 21.2 \\
\hspace{0.4cm}w/ highlight on both input image and evidence & & \textbf{31.6} & 29.3 & & 23.0 & 22.4 & 22.7 \\
\rowcolor{OurColor}
\hspace{0.4cm}\textbf{\ours (Visual)} -- highlight on evidence only & & \textbf{31.6} & \textbf{29.4} & & \textbf{24.1} & \textbf{23.7} & \textbf{23.9} \\
\midrule
\rowcolor{TitleColor}
$\blacktriangledown$ Qwen2.5-VL-7B (only input image) & & 32.1 & 30.2 & & 23.9 & 25.1 & 24.5 \\
\hspace{0.4cm}w/ bbox on input image & & 32.0 & 29.9 & & 23.5 & 24.9 & 24.2 \\
\hspace{0.4cm}w/ only selected evidence & & 32.1 & 29.9 & & 23.9 & 25.1 & 24.5 \\
\hspace{0.4cm}w/ input image and evidence (no highlight) & & 30.7 & 28.4 & & 23.2 & 24.5 & 23.8 \\
\hspace{0.4cm}w/ highlight on both input image and evidence & & 31.7 & 29.6 & & 22.5 & 25.4 & 24.0 \\
\rowcolor{OurColor}
\hspace{0.4cm}\textbf{\ours (Visual)} -- highlight on evidence only & & \textbf{32.4} & \textbf{30.2} & & \textbf{24.4} & \textbf{25.4} & \textbf{24.9} \\
\bottomrule
\end{tabular}
}
\vspace{-0.15cm}
\caption{Ablation study on visual highlighting strategies.}
\label{tab:ablation2}
\end{table} 

\begin{table}[t]
    \centering
\small
\setlength{\tabcolsep}{.2em}
\resizebox{\linewidth}{!}{
\begin{tabular}{lc cc c ccc c}
\toprule
& & \multicolumn{2}{c}{\textbf{E-VQA}} & & \multicolumn{3}{c}{\textbf{InfoSeek}} \\
\cmidrule{3-4} \cmidrule{6-8}
 & & Single & All & & U-Q & U-E & All \\
\midrule
\rowcolor{TitleColor}
$\blacktriangledown$ Qwen2.5-VL-3B & & 30.3 & 28.0 & & 22.6 & 22.2 & 22.4 \\
\hspace{0.4cm}w/ highlight on all retrieved content & & 30.6 & 28.2 & & 24.1 & 23.5 & 23.8 \\
\hspace{0.4cm}w/ highlight on entire passage & & 31.2 & 28.8 & & 25.0 & 23.9 & 24.4 \\
\hspace{0.4cm}w/ highlight on evidence only (no discard) & & \textbf{31.7} & \textbf{29.1} & & 24.5 & 24.0 & 24.2 \\
\hspace{0.4cm}w/ keep only evidence & & 30.5 & 28.5 & & 27.0 & 26.3 & 26.7 \\
\rowcolor{OurColor}
\hspace{0.4cm}\textbf{\ours (Textual)} & & \textbf{31.7} & \textbf{29.1} & & \textbf{29.8} & \textbf{29.2} & \textbf{29.5} \\
\midrule
\rowcolor{TitleColor}
$\blacktriangledown$ Qwen2.5-VL-7B & & 32.1 & 30.2 & & 23.9 & 25.1 & 24.5 \\
\hspace{0.4cm}w/ highlight on all retrieved content & & 32.0 & 30.1 & & 24.1 & 25.0 & 24.5 \\
\hspace{0.4cm}w/ highlight on entire passage & & 32.1 & 30.0 & & 26.0 & 25.9 & 26.0 \\
\hspace{0.4cm}w/ highlight on evidence only (no discard) & & 33.6 & 31.4 & & 24.9 & 25.4 & 25.1 \\
\hspace{0.4cm}w/ keep only evidence & & 33.9 & 31.6 & & 26.3 & 26.4 & 26.4 \\
\rowcolor{OurColor}
\hspace{0.4cm}\textbf{\ours (Textual)} & & \textbf{34.7} & \textbf{32.1} & & \textbf{28.8} & \textbf{29.9} & \textbf{29.4} \\
\bottomrule
\end{tabular}
}
\vspace{-0.15cm}
\caption{Ablation study on textual highlighting strategies.}
\label{tab:ablation3}
\end{table}

\tit{Comparison with Different Textual Highlighting Strategies}
Table~\ref{tab:ablation3} compares different uses of the textual relevance
scores produced by \ours. Beyond the unmodified baseline, we consider
highlighting the complete retrieved context
(\textit{w/ highlight on all retrieved content}) or the full passages
containing the selected evidence
(\textit{w/ highlight on entire passage}). At sentence level,
\textit{w/ highlight on evidence only (no discard)} marks sentences whose
scores exceed $\alpha$ while retaining the remaining context, whereas
\textit{w/ keep only evidence} removes all non-selected sentences,
equivalently setting $\alpha_{\mathrm{drop}}=\alpha$. The complete \ours
strategy highlights high-relevance sentences, discards low-relevance ones,
and retains intermediate-scoring sentences unchanged.

Highlighting the complete retrieved context provides only marginal benefits,
whereas more fine-grained sentence-level strategies are generally more
effective than passage-level highlighting. Retaining only the selected
evidence substantially improves InfoSeek, but can remove supporting context
that remains useful, particularly on E-VQA. The complete \ours strategy
achieves the best performance across all model and dataset configurations,
showing that explicit highlighting and selective filtering play complementary
roles.

\begin{table}[t]
    \centering
\small
\setlength{\tabcolsep}{.18em}
\resizebox{\linewidth}{!}{
\begin{tabular}{l cccc c c c}
\toprule
 & RealWorldQA & V-Star & TextVQA & ChartQA & & \textbf{Avg} & \\
\midrule

$\blacktriangledown$ Qwen2.5-VL-3B
& 59.1 & 66.5 & 62.5 & 79.1
& & 66.8 & \\

\rowcolor{OurColor}
\hspace{0.4cm}\textbf{+ \ours (Ours)}
& \textbf{61.7} & \textbf{71.2} & \textbf{66.4} & \textbf{79.5}
& & \textbf{69.7} & \inc{2.9} \\

\midrule

$\blacktriangledown$ Qwen3-VL-4B
& 66.7 & 70.2 & 74.7 & 80.7
& & 73.1 & \\

\rowcolor{OurColor}
\hspace{0.4cm}\textbf{+ \ours (Ours)}
& \textbf{71.6} & \textbf{78.5} & \textbf{76.5} & \textbf{82.3}
& & \textbf{77.2} & \inc{4.2} \\

\midrule

$\blacktriangledown$ Qwen2.5-VL-7B
& 65.0 & 71.2 & 75.7 & 77.0
& & 72.2 & \\

\rowcolor{OurColor}
\hspace{0.4cm}\textbf{+ \ours (Ours)}
& \textbf{67.5} & \textbf{73.3} & \textbf{77.9} & \textbf{79.4}
& & \textbf{74.5} & \inc{2.3} \\

\midrule

$\blacktriangledown$ Qwen3-VL-8B
& 66.8 & 73.3 & 76.9 & 82.0
& & 74.8 & \\

\rowcolor{OurColor}
\hspace{0.4cm}\textbf{+ \ours (Ours)}
& \textbf{69.7} & \textbf{77.0} & \textbf{77.2} & \textbf{82.8}
& & \textbf{76.7} & \inc{1.9} \\

\midrule

$\blacktriangledown$ Qwen2.5-VL-32B
& 65.6 & \textbf{73.8} & 72.4 & 79.7
& & 72.9 & \\

\rowcolor{OurColor}
\hspace{0.4cm}\textbf{+ \ours (Ours)}
& \textbf{67.1} & 72.3 & \textbf{74.0} & \textbf{82.2}
& & \textbf{73.9} & \inc{1.0} \\

\bottomrule
\end{tabular}
}
\vspace{-0.15cm}
\caption{Performance comparison on standard MLLM benchmarks. For each
backbone, we compare the original model with the same model augmented with
\ours using only visual highlighting. Results are reported on RealWorldQA,
V-Star, TextVQA, and ChartQA. Avg denotes the average score across the four
benchmarks; improvements over the baseline are shown in green.}
\label{tab:mllm_benchmarks}
\end{table}

\subsection{Results on Standard MLLM Benchmarks}
To further assess the robustness of our visual-highlighting strategy, we evaluate \ours on standard MLLM benchmarks beyond KB-VQA. Specifically, we report results on RealWorldQA~\cite{grok}, V-Star~\cite{wu2024v}, TextVQA~\cite{singh2019towards}, and ChartQA~\cite{masry2022chartqa}, covering real-world visual reasoning, scene-text understanding, and chart reasoning. Since these benchmarks do not provide retrieved textual context, we apply \ours using visual highlighting only. 
Moreover, as several questions require global scene understanding or relationships between multiple objects, we retain the full input image and append the selected crop enclosed by the visual highlighting markers. This preserves the complete visual context while explicitly directing the model toward the most query-relevant region.

As shown in Table~\ref{tab:mllm_benchmarks}, \ours improves the average performance of all five evaluated backbones, with gains ranging from $+1.0$ to $+4.2$ points. These results show that highlighting query-relevant image regions improves visual evidence utilization even without retrieved textual context, extending the applicability of \ours beyond KB-VQA.

\section{Additional Implementation Details}

\subsection{Pre-Processing and Model-Specific Details}

\tinytit{Target Object Identification}
As described in the main paper, visual evidence highlighting requires identifying the target object referred to in the question. We denote by $\mathcal{T_\text{obj}}$ the set of indices of the input tokens corresponding to the queried object. These indices are then used to select the row indices of the submatrix $\mathbf{A}^{\ell,k}_{\text{obj}\rightarrow\text{vis}}$.
To extract $\mathcal{T_\text{obj}}$, we employ the spaCy NLP toolkit\footnote{Specifically, we use the \texttt{en\_core\_web\_sm} model available at \url{https://spacy.io/models/en}.}, leveraging dependency parsing and part-of-speech annotations through a lightweight, linguistically grounded procedure. 

Given an input question, we analyze its syntactic structure to identify the noun phrase that captures the focus of the question. In particular, we consider nouns associated with interrogative constructions (\eg, \textit{what}, \textit{which}, \textit{who}) as well as nouns involved in relevant syntactic relations, such as verbal or prepositional dependencies. Once a candidate head noun is identified, we expand it to the corresponding full noun phrase by including syntactically attached modifiers, such as compound nouns and adjectival modifiers, obtaining a semantically coherent span that represents the queried object~\footnote{For example, in the sentence ``\textit{John James Audubon became famous for painting what?}'', \textit{John James Audubon} is selected as the visual subject.}.
If this procedure does not identify a valid target-object span, we fall back to using the last token in input (\ie, the same token used for text highlighting) as $\mathcal{T_\text{obj}}$. This fallback ensures that visual-evidence extraction remains well-defined even for questions whose syntactic structure does not contain an explicit or reliably detectable target object\footnote{For example, ``\textit{What is green?}'' contains no noun phrase denoting the object that should be localized.}.

\begin{table}[t]
    \centering
    \small
    \setlength{\tabcolsep}{0.75em}
    \resizebox{0.95\linewidth}{!}{
    \begin{tabular}{lrrr}
        \toprule
        \textbf{Dataset} & Questions & Entity Identified & Fallback \\
        \midrule
        E-VQA    & 5,750  & 5,746 (99.93\%) & 4 (0.07\%) \\
        InfoSeek & 71,335 & 71,335 (100.00\%) & 0 (0.00\%) \\
        OVEN     & 3,291  & 2,995 (91.01\%) & 296 (8.99\%) \\
        ViQuAE   & 1,257  & 1,232 (98.01\%) & 25 (1.99\%) \\
        \midrule
        Overall  & 81,633 & 81,308 (99.60\%) & 325 (0.40\%) \\
        \bottomrule
    \end{tabular}
    }
    \vspace{-0.15cm}
    \caption{Coverage of the dataset-specific entity-selection procedure. A fallback occurs when no valid entity is identified and the last input token is used to extract the attention map.}
    \label{tab:entity_selection_coverage}
\end{table}

To validate this step, we measure how often the entity-selection procedure identifies a valid textual target for extracting the visual attention map. The results in Table~\ref{tab:entity_selection_coverage} reflect the different linguistic characteristics of the four datasets. E-VQA and InfoSeek mostly contain templated questions in which the visual entity is explicitly introduced by demonstrative expressions such as \emph{this}, \emph{that}, \emph{these}, or \emph{those}. In these cases, the nominal expression associated with the demonstrative provides a reliable attention target. Consequently, the procedure identifies an entity in 99.93\% of E-VQA questions and all InfoSeek questions.
OVEN contains less regular, often short or underspecified questions. Nevertheless, a valid noun phrase is identified in 2,995 of the 3,291 questions (91.01\%). The remaining 296 questions often describe an attribute without explicitly naming the corresponding object.
ViQuAE instead contains longer, more compositional questions that may mention multiple entities. We therefore use named-entity recognition and select the last named entity detected in the question, which generally corresponds to the entity closest to the requested information. This strategy identifies a valid target in 1,232 of the 1,257 questions (98.01\%), with the last-token fallback used for the remaining 25.

\begin{table}[t]
    \centering
    \small
    \renewcommand{\arraystretch}{0.85}
    \resizebox{0.75\linewidth}{!}{
        \begin{tabular}{ll}
        \toprule
        \textbf{Model} & \textbf{Sink Dimensions} \\
        \midrule
        Qwen2-VL-2B-Instruct    & 1073, 534, 940 \\
        Qwen2-VL-7B-Instruct    & 2570, 458 \\
        \midrule
        Qwen2.5-VL-3B-Instruct  & 318, 1874, 1819 \\
        Qwen2.5-VL-7B-Instruct  & 458, 2570 \\
        Qwen2.5-VL-32B-Instruct & 4675, 3094 \\
        \midrule
        Qwen3-VL-4B-Instruct    & 0 \\
        Qwen3-VL-8B-Instruct    & 1838 \\
        \midrule
        InternVL3.5-4B          & 4, 396, 0 \\
        InternVL3.5-8B          & 2276, 233 \\
        InternVL3.5-38B         & 731 \\

        \bottomrule
    \end{tabular}
    }
    \vspace{-0.15cm}
    \caption{Attention-sink dimensions filtered for each MLLM.}
    \label{tab:model_sink_dimensions}
\end{table}

\tit{Context Sentence Splitting}
To obtain sentence-level relevance scores (\eg, $\mathbf{\hat{a}}_{\mathrm{txt}}$) from the token-level attention vector $\mathbf{a}_{\mathrm{txt}}$, we first segment the retrieved context into sentences using the English spaCy pipeline, which is also employed for target-object extraction. Specifically, given a context $\mathbf{X}_C$, we obtain its sentences as
$[
\mathcal{S}(\mathbf{X}_C)={s_1,\ldots,s_M}.
]$
For each sentence $s_i$, we identify the corresponding context tokens and compute its relevance score by averaging their token-level attention values. The resulting sentence-level scores are then used to determine which sentences should be highlighted, retained as supporting context, or discarded.

\tit{Model Sink Dimensions}
Table~\ref{tab:model_sink_dimensions} reports the hidden dimensions identified as attention sinks and filtered for each evaluated model. To determine them, we process 100 randomly sampled examples from the E-VQA validation split and rank the dimensions of the \texttt{BOS} hidden representation according to their average activation magnitude. We then select the dimensions that consistently exhibit the highest values across the analyzed samples and keep them fixed in all subsequent experiments.

Under standard causal self-attention, the \texttt{BOS} token cannot attend to any subsequent token. Therefore, as long as it always occupies the first sequence position and is encoded identically, its hidden representation is expected to be largely input-independent, making the associated sink dimensions stable across examples. Evaluating multiple samples serves primarily to verify this stability empirically and to reduce the influence of model-specific numerical effects or input-formatting artifacts on the selected dimensions.

\subsection{Training-Free Baselines Implementation}
We evaluate the compatibility of \ours with reasoning, passage-filtering, and re-ranking strategies. In each case, the underlying method first processes the retrieved context according to its original design, after which \ours refines the resulting visual and textual evidence before answer generation.

\tit{Chain-of-Thought Prompting~\cite{wei2022chain}}
We augment the standard generation prompt by instructing the MLLM to reason step by step over the image and retrieved context before producing the final answer. The backbone, retrieved documents, and decoding settings remain unchanged, requiring neither training nor additional models.

\tit{Prompt-based Critic~\cite{compagnoni2025reag}}
Each passage from the top-$3$ retrieved documents is independently evaluated by the same zero-shot MLLM used for answer generation, which is prompted to classify it as useful or irrelevant given the image and question following the original prompt used in~\cite{compagnoni2025reag}. Only passages classified as useful are retained and provided to the final generation step.

\tit{Trained Critic~\cite{compagnoni2025reag}}
We also employ the official Critic model proposed in~\cite{compagnoni2025reag}, based on Qwen2.5-VL-3B and fine-tuned for passage-relevance classification. Following its original configuration, the Critic model filters passages from the top-$20$ retrieved documents before they are provided to the generator.

\tit{EchoSight Re-ranker ~\cite{yan2024echosight}}
We divide the top-$3$ retrieved documents into sections and adapt the trained EchoSight Q-Former re-ranker to score each candidate section according to its relevance to the image-question pair. Unlike the complete EchoSight pipeline, we retain our original retrieval results and use only its multimodal re-ranking component. The sections are then ordered according to their relevance scores, and the top-$5$ are provided to the answer generator. This configuration keeps
the initial retrieval pipeline and the final generator fixed, isolating the effect of the re-ranking component.

\tit{Qwen3-VL-8B-Reranker~\cite{li2026qwen3}}
Following the same section-level setting, we use the released
Qwen3-VL-8B-Reranker to estimate the relevance of each candidate section to the multimodal image-question query. The sections are ordered according to the resulting relevance scores, and the top-$5$ are provided to the answer generator. The initial retrieval pipeline and the final generator remain
unchanged.

\subsection{Computing Infrastructure}
Experiments were conducted on compute nodes equipped with four NVIDIA A100 GPUs with 64\,GiB of HBM2e memory each, one 32-core Intel Xeon Platinum 8358 CPU at 2.6\,GHz, and 512\,GiB of DDR4 system memory. All model-inference jobs were restricted to a single compute node. The standard allocation comprised one GPU, 8 CPU cores, and 128\,GiB of system memory. Depending on the memory requirements of the backbone, individual jobs used between one and four GPUs. The largest model configurations used all four GPUs, corresponding to 256\,GiB capacity of GPU memory. Dataset shards were evaluated independently using Slurm array jobs; no multi-node model parallelism was employed.

The experiments were run with Python~3.10.19, PyTorch~2.6.0 (CUDA~12.6 build), Transformers~4.57.1, Accelerate~1.12.0, FlashAttention-2~2.8.3, and FAISS-GPU~1.7.2. Image processing used Pillow~10.3.0 and OpenCV~4.11.0, while spaCy~3.7.5 was used for the target-object extraction procedure. Since \ours is training-free, the reported computation concerns
retrieval, inference, and evaluation only; no model parameters were optimized.
\section{Hyperparameter Analysis and Discussion}

\subsection{Textual Evidence Selection}
\tinytit{Selection of $\alpha$ and $\alpha_{\mathrm{drop}}$}
We select $\alpha$ on the E-VQA validation split. Sentence precision measures the fraction of highlighted sentences found in the source document, whereas document recall measures the fraction of retrieval-hit samples containing at least one such sentence. As shown in Table~\ref{tab:alpha_selection}, $\alpha=0.5$ provides the best trade-off: it preserves substantially higher recall than $\alpha=0.75$, while improving precision and eliciting considerably fewer sentences than $\alpha=0.25$. It also achieves the highest end-to-end validation accuracy and is therefore used in all subsequent experiments.

\begin{table}[t]
    \centering
    \small
    \setlength{\tabcolsep}{0.65em}
    \resizebox{0.9\linewidth}{!}{
    \begin{tabular}{ccccc}
        \toprule
        $\alpha$ & Accuracy & Precision & Recall & Avg. Sentences \\
        \midrule
        0.25 & 32.14 & 57.72 & 93.89 & 5.86 \\
        \rowcolor{OurColor}
        \textbf{0.50} & \textbf{32.61} & 62.07 & 77.62 & 2.03 \\
        0.75 & 31.36 & 68.97 & 60.44 & 1.04 \\
        \bottomrule
    \end{tabular}
    }
    \vspace{-0.15cm}
    \caption{Selection of $\alpha$ on the E-VQA validation split using Qwen2.5-VL-3B.}
        \label{tab:alpha_selection}
    \vspace{-.2cm}
\end{table}

\begin{table}[t]
    \centering
    \small
    \resizebox{0.68\linewidth}{!}{
    \begin{tabular}{ccc}
        \toprule
        $\alpha_{\mathrm{drop}}$
        & Accuracy
        & Context reduction \\
        \midrule
        $\alpha / 2.0$ & 31.59 & 74.47\% \\
        \rowcolor{OurColor}
        $\mathbf{\alpha / 3.0}$ & \textbf{32.61} & 70.45\% \\
        $\alpha / 4.0$ & 31.92 & 66.71\% \\
        \bottomrule
    \end{tabular}
    }
    \vspace{-0.15cm}
    \caption{Selection of $\alpha_{\mathrm{drop}}$ on the E-VQA validation split using Qwen2.5-VL-3B.}
    \label{tab:alpha_drop_selection}
\end{table}

\tit{Selection of $\alpha_{\mathrm{drop}}$}
We select the sentence-dropping threshold on the E-VQA validation split, by evaluating $\alpha_{\mathrm{drop}}\in\{\alpha / 2,\alpha / 3,\alpha / 4\}$. As shown in Table~\ref{tab:alpha_drop_selection}, the most aggressive setting reduces the context by $74.47\%$, but yields lower accuracy, while retaining more context with $\alpha_{\mathrm{drop}}=\alpha / 4$ does not improve performance. We therefore adopt $\alpha_{\mathrm{drop}}=\alpha / 3$, which achieves the highest validation accuracy while reducing the retrieved context by $70.45\%$. For InternVL3.5, we note that sentence-level attention distributions are more sharply peaked, causing the default threshold $\alpha/3$ to discard approximately $84\%$ of the context. We therefore use $\alpha/6$, reducing the discarded fraction to approximately $70\%$ and aligning its filtering behavior with the other model families.

\begin{figure}[t]
\centering
\scriptsize
\setlength{\tabcolsep}{0.1em}
\resizebox{\linewidth}{!}{
\begin{tabular}{ccc c ccc}
\textbf{Original} & \textbf{Attention Map} & \textbf{Filtered} & & 
\textbf{Original} & \textbf{Attention Map} & \textbf{Filtered} \\[-2pt]
\textbf{Image} & \textbf{(No Filtering)} & \textbf{Attention Map} & & 
\textbf{Image} & \textbf{(No Filtering)} & \textbf{Attention Map} \\[3pt]

\includegraphics[width=0.19\linewidth, height=0.12\linewidth ]{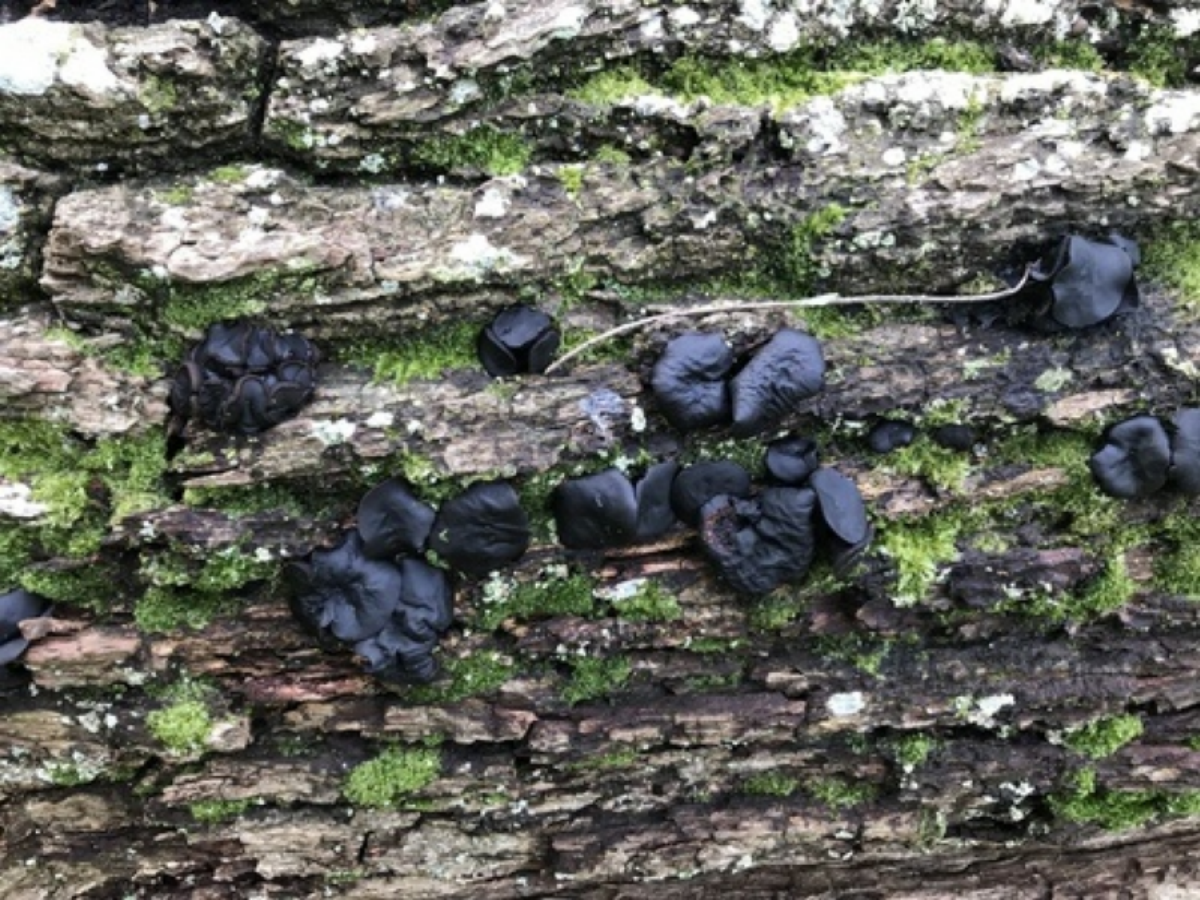} &
\includegraphics[width=0.19\linewidth, height=0.12\linewidth ]{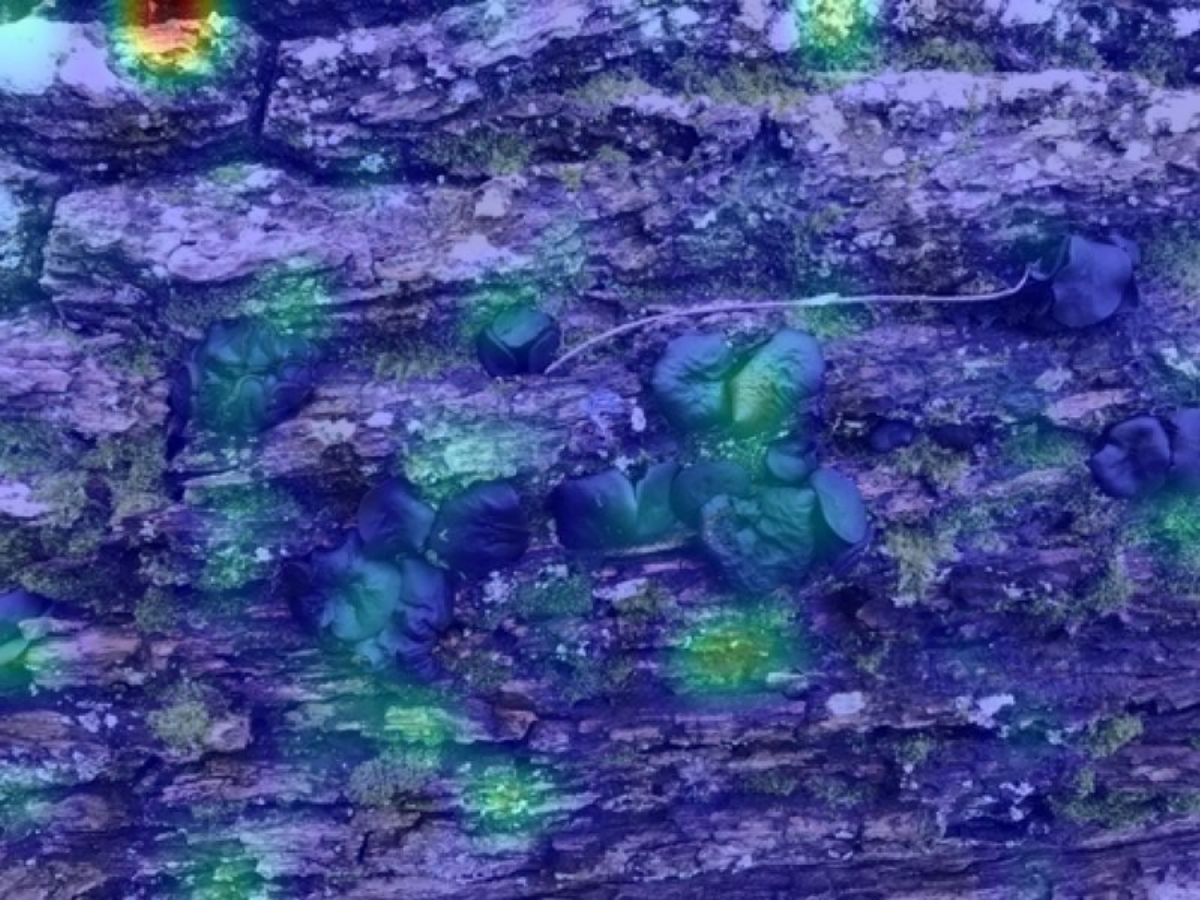} &
\includegraphics[width=0.19\linewidth, height=0.12\linewidth ]{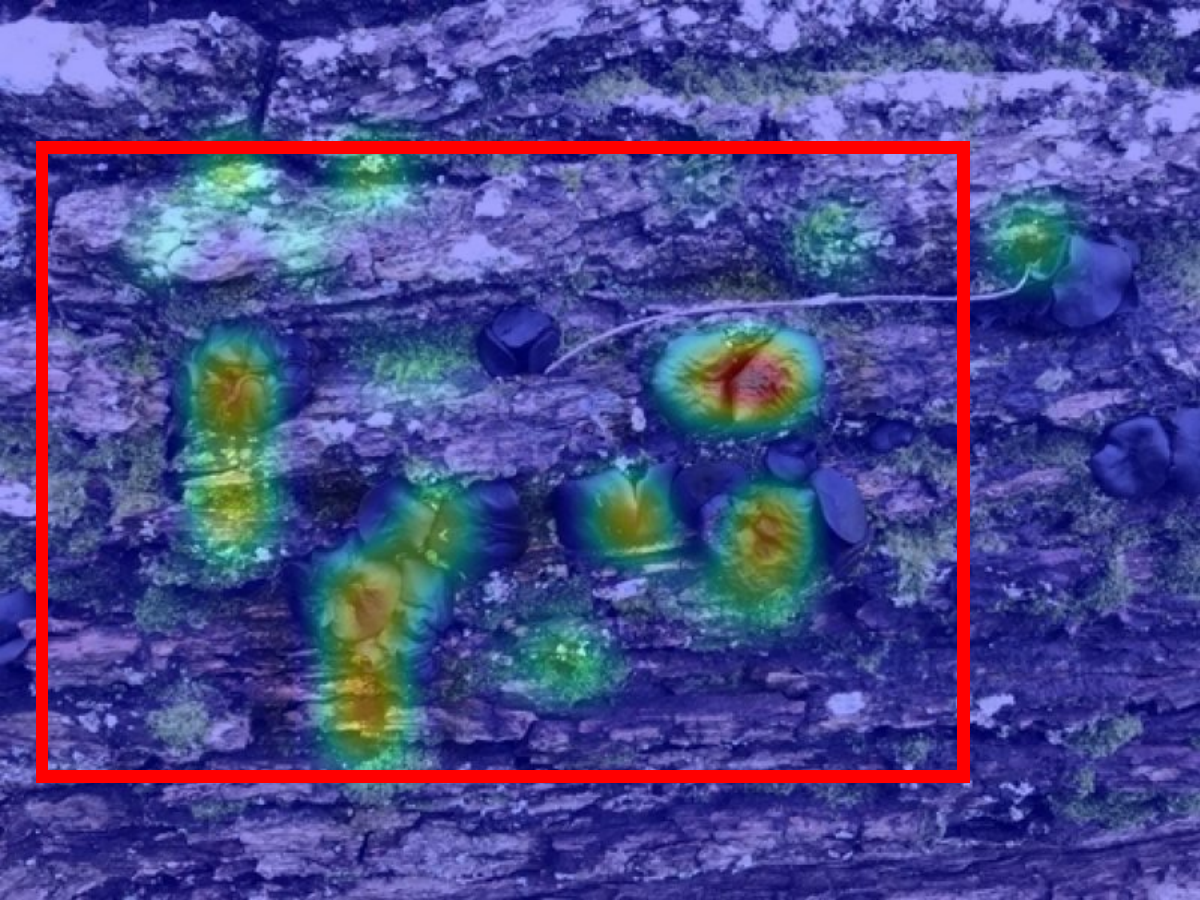} & &
\includegraphics[width=0.19\linewidth, height=0.12\linewidth ]{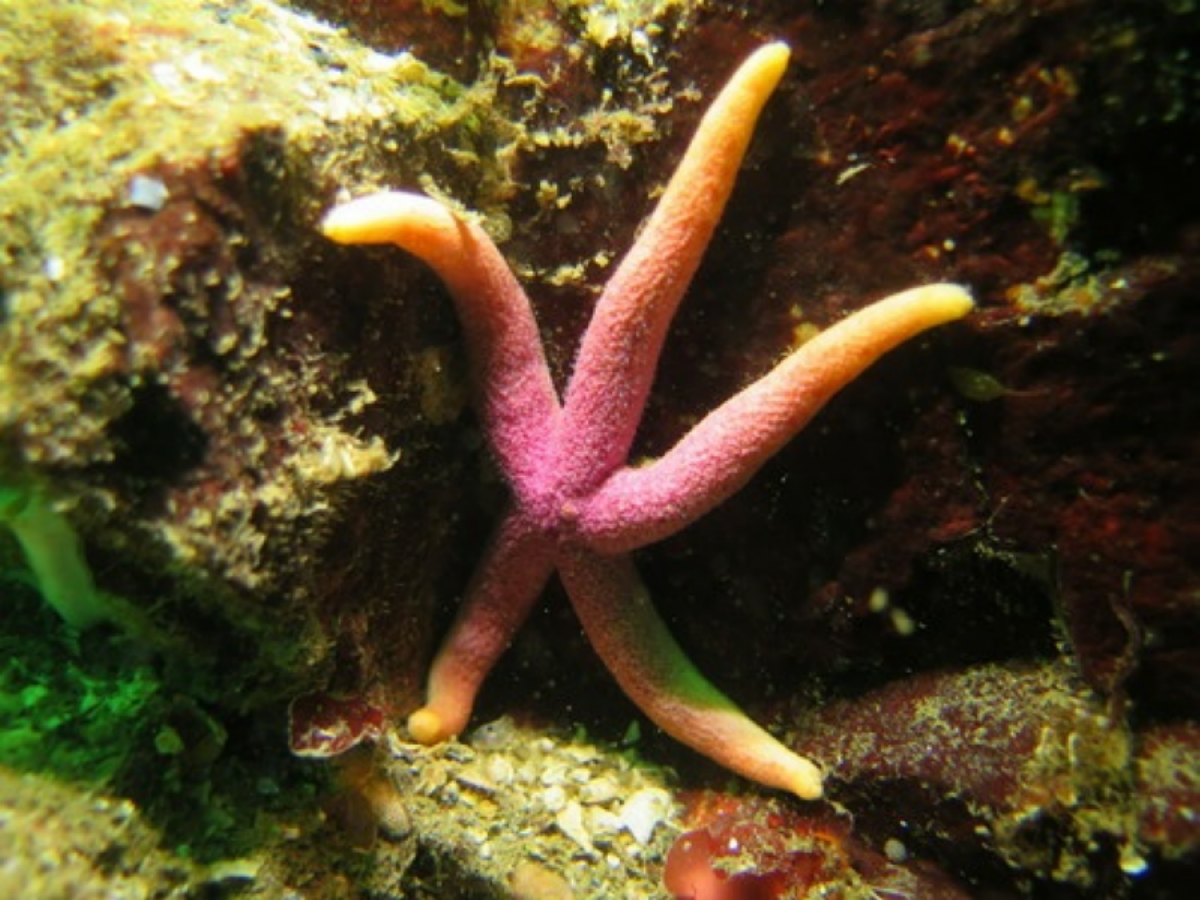} &
\includegraphics[width=0.19\linewidth, height=0.12\linewidth ]{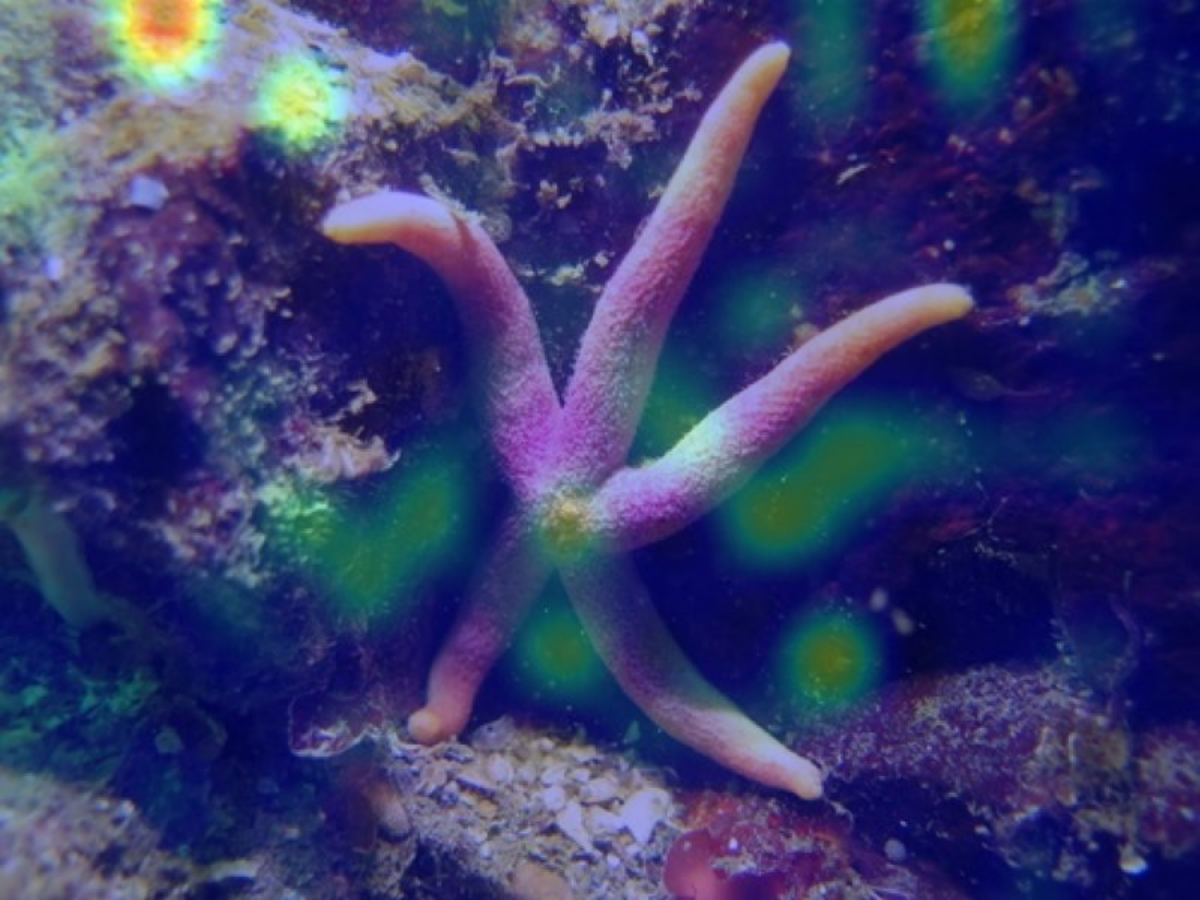} &
\includegraphics[width=0.19\linewidth, height=0.12\linewidth ]{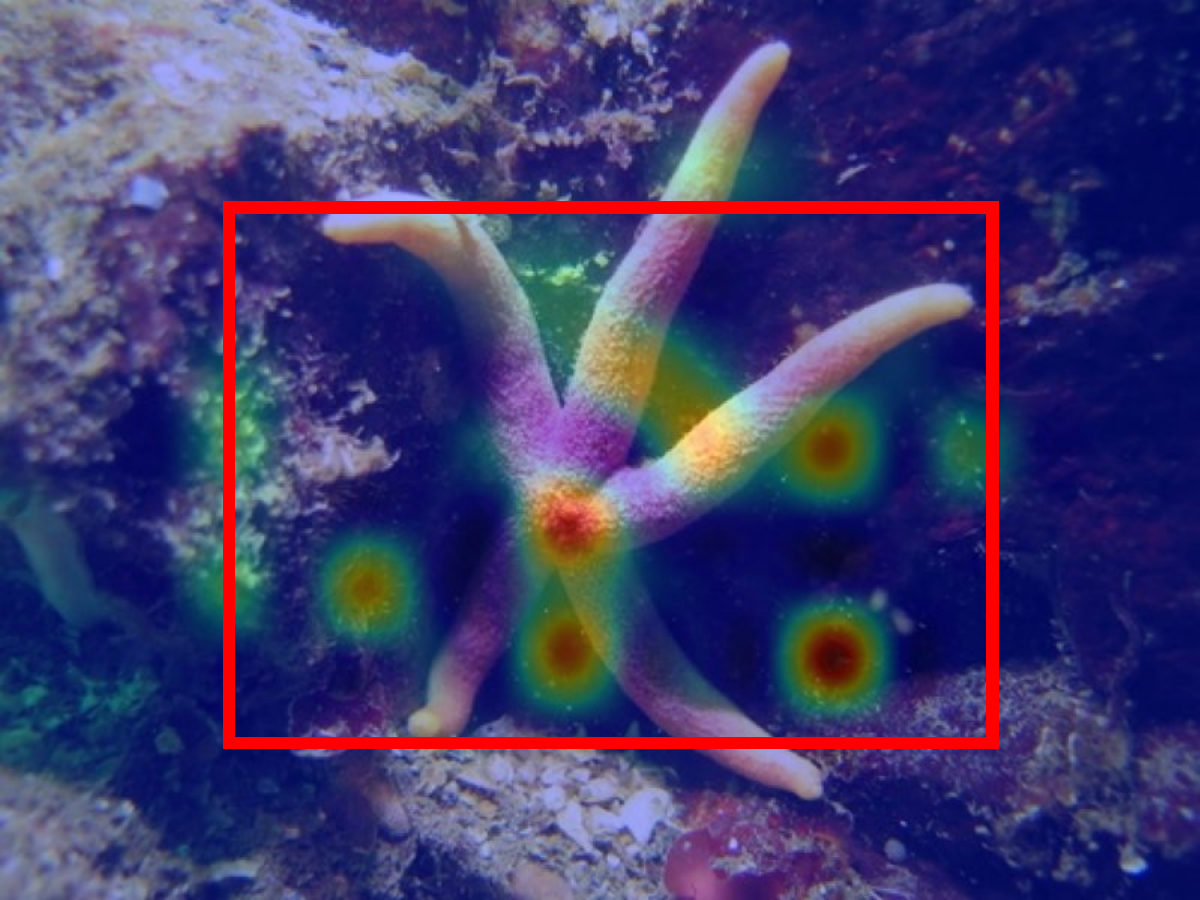} \\

\includegraphics[width=0.19\linewidth, height=0.12\linewidth ]{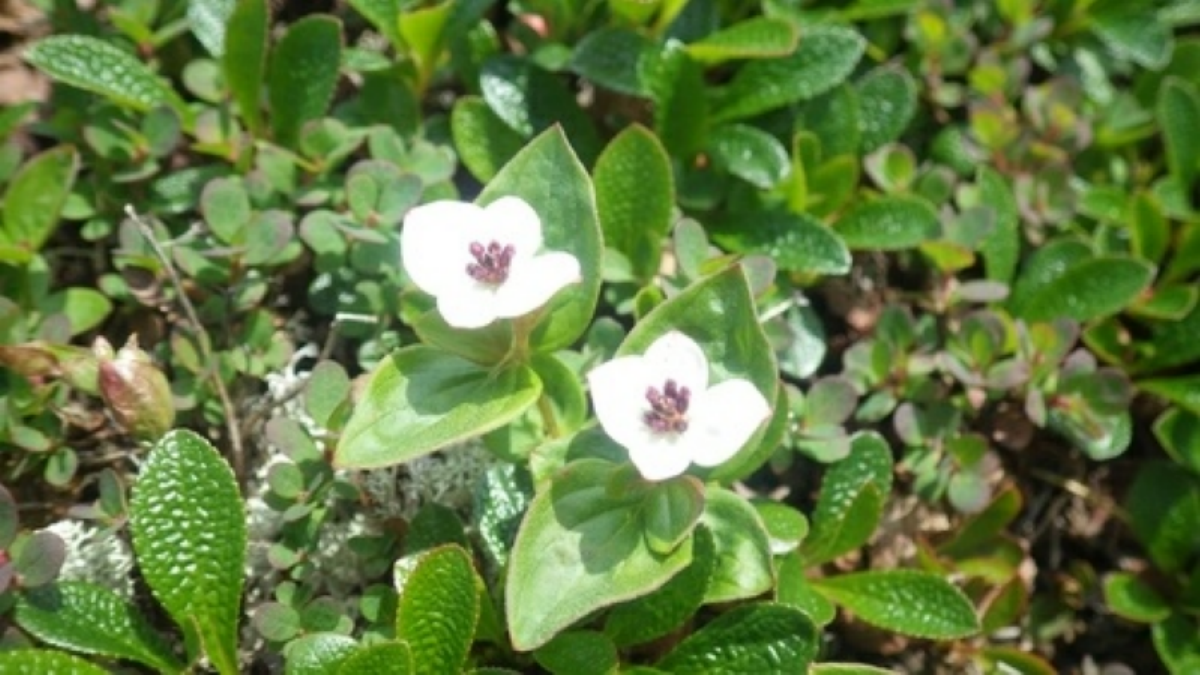} &
\includegraphics[width=0.19\linewidth, height=0.12\linewidth ]{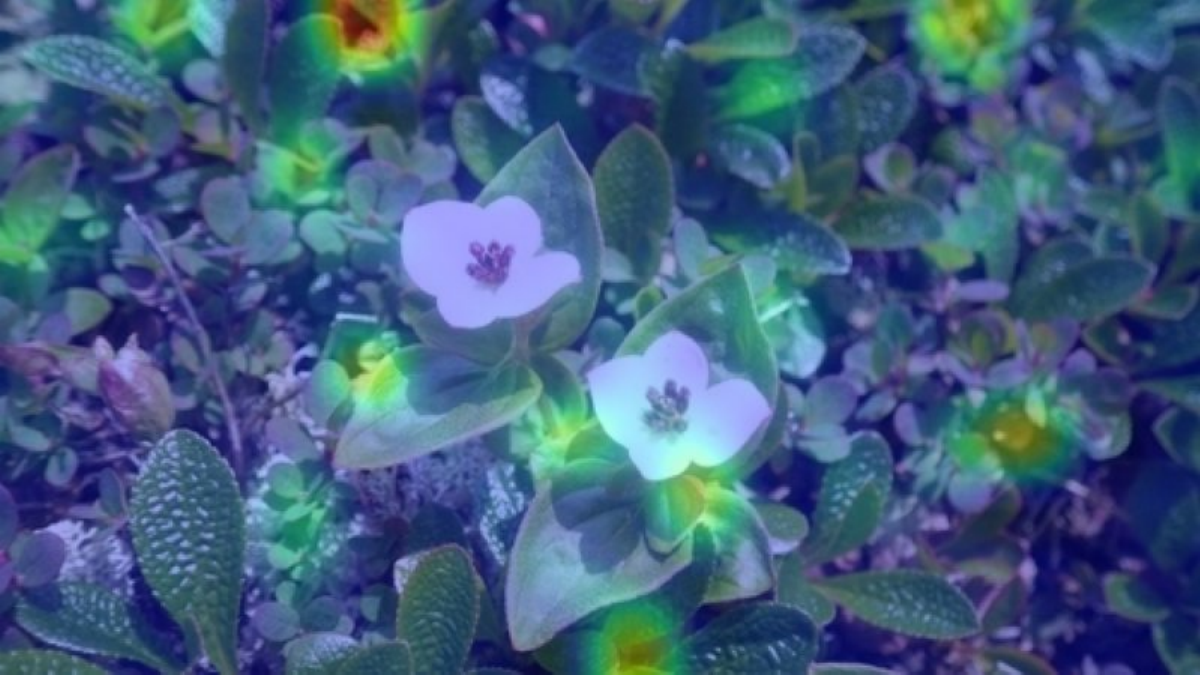} &
\includegraphics[width=0.19\linewidth, height=0.12\linewidth ]{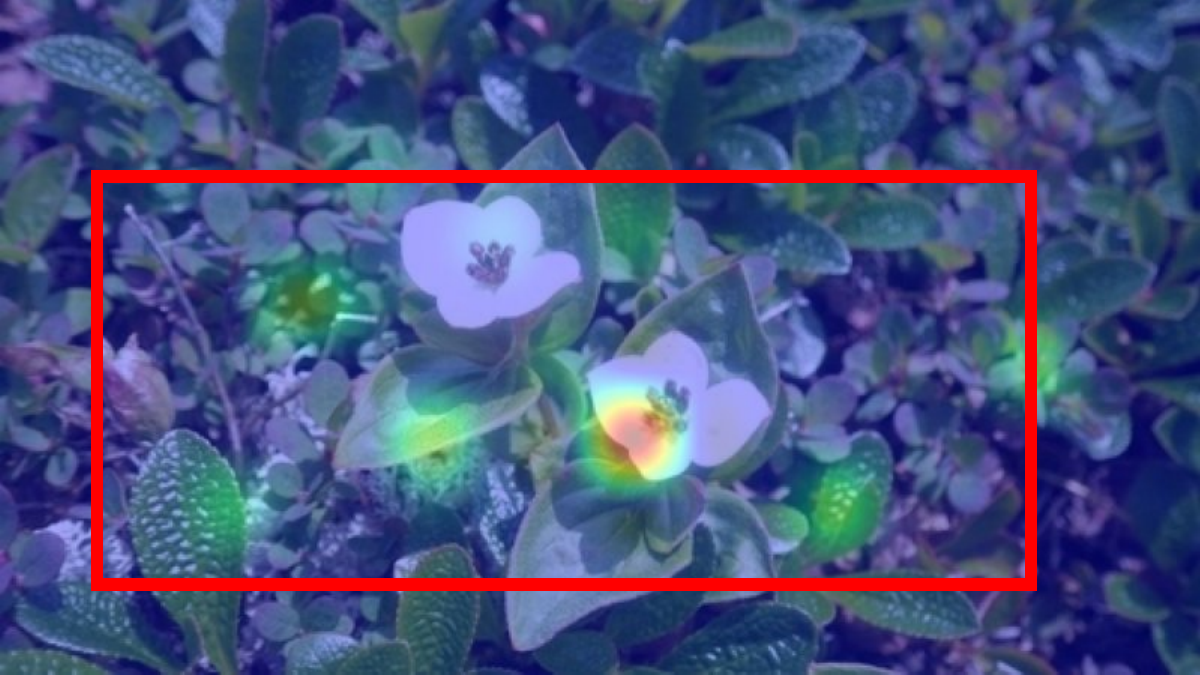} & &
\includegraphics[width=0.19\linewidth, height=0.12\linewidth ]{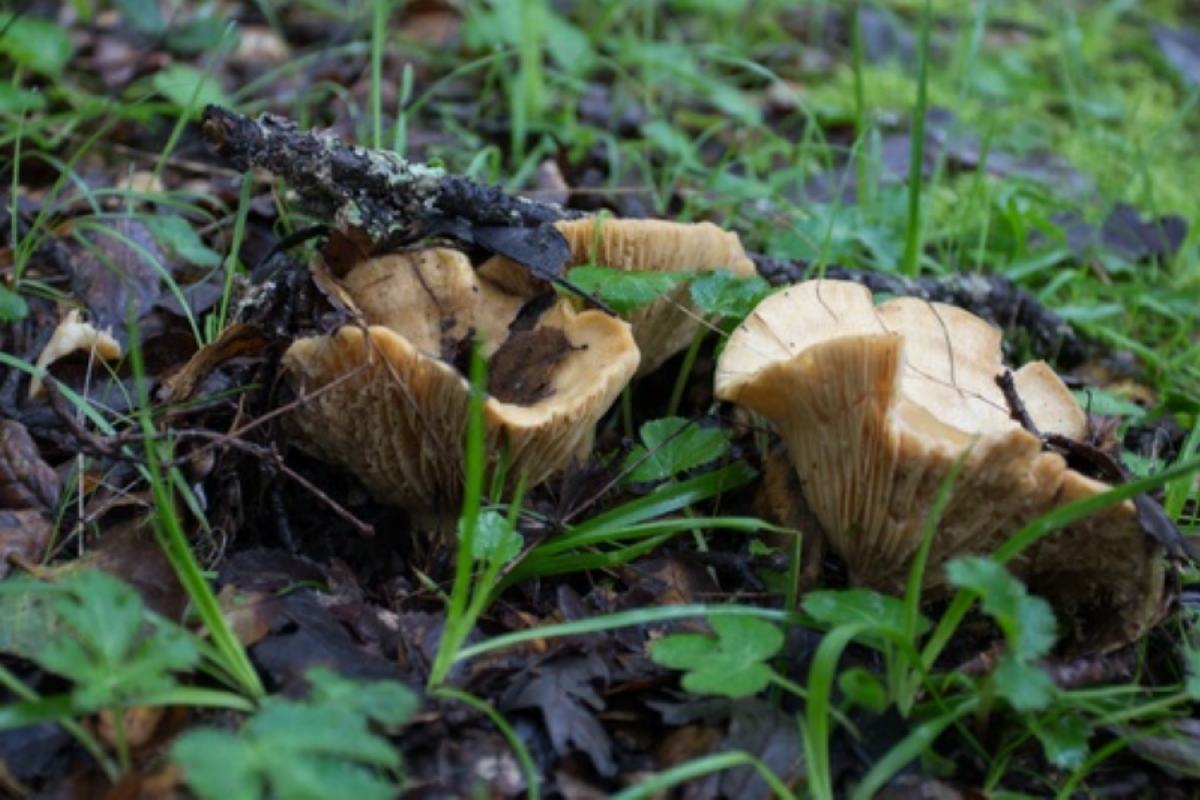} &
\includegraphics[width=0.19\linewidth, height=0.12\linewidth ]{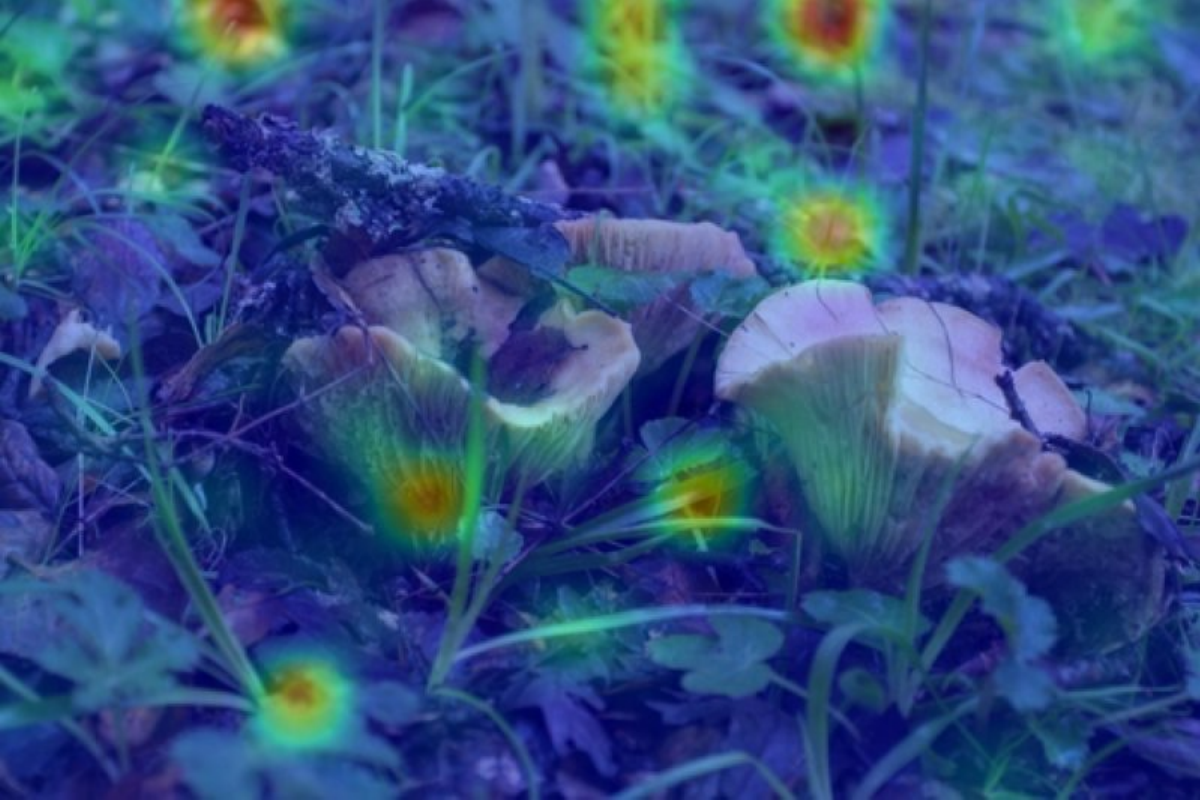} &
\includegraphics[width=0.19\linewidth, height=0.12\linewidth ]{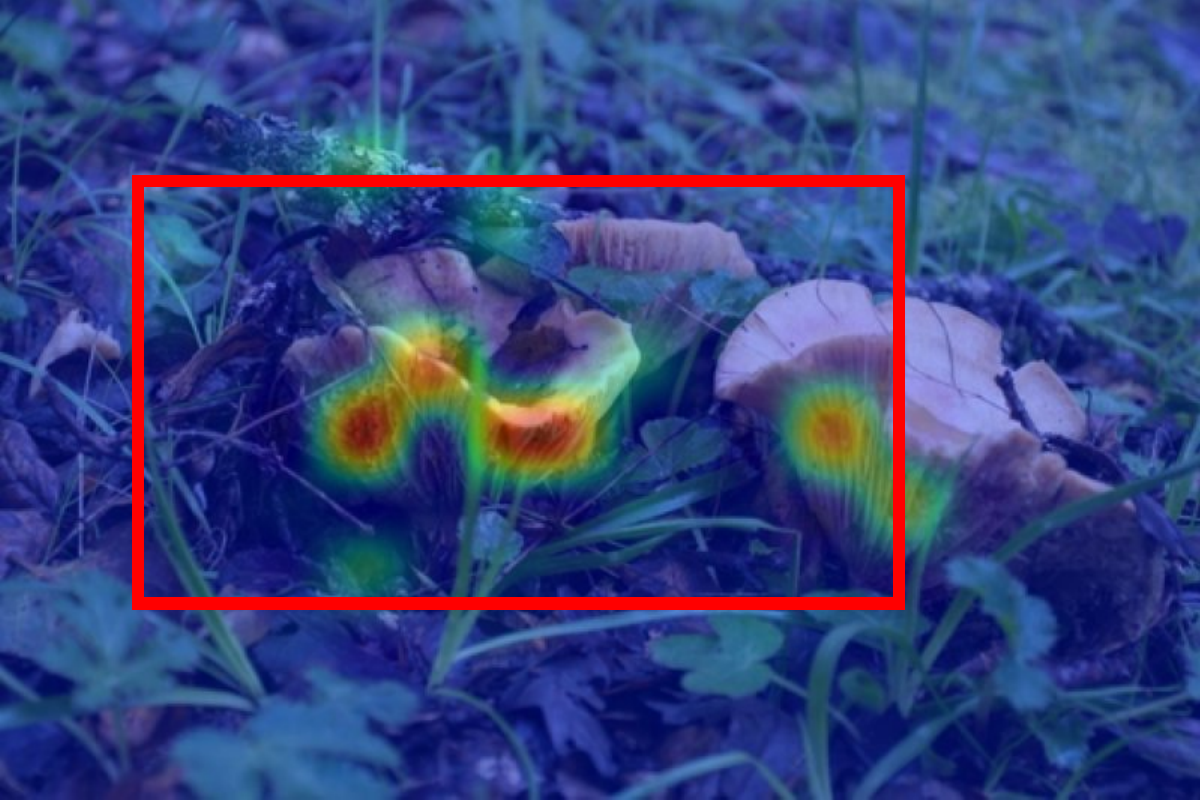} \\

\includegraphics[width=0.19\linewidth, height=0.24\linewidth ]{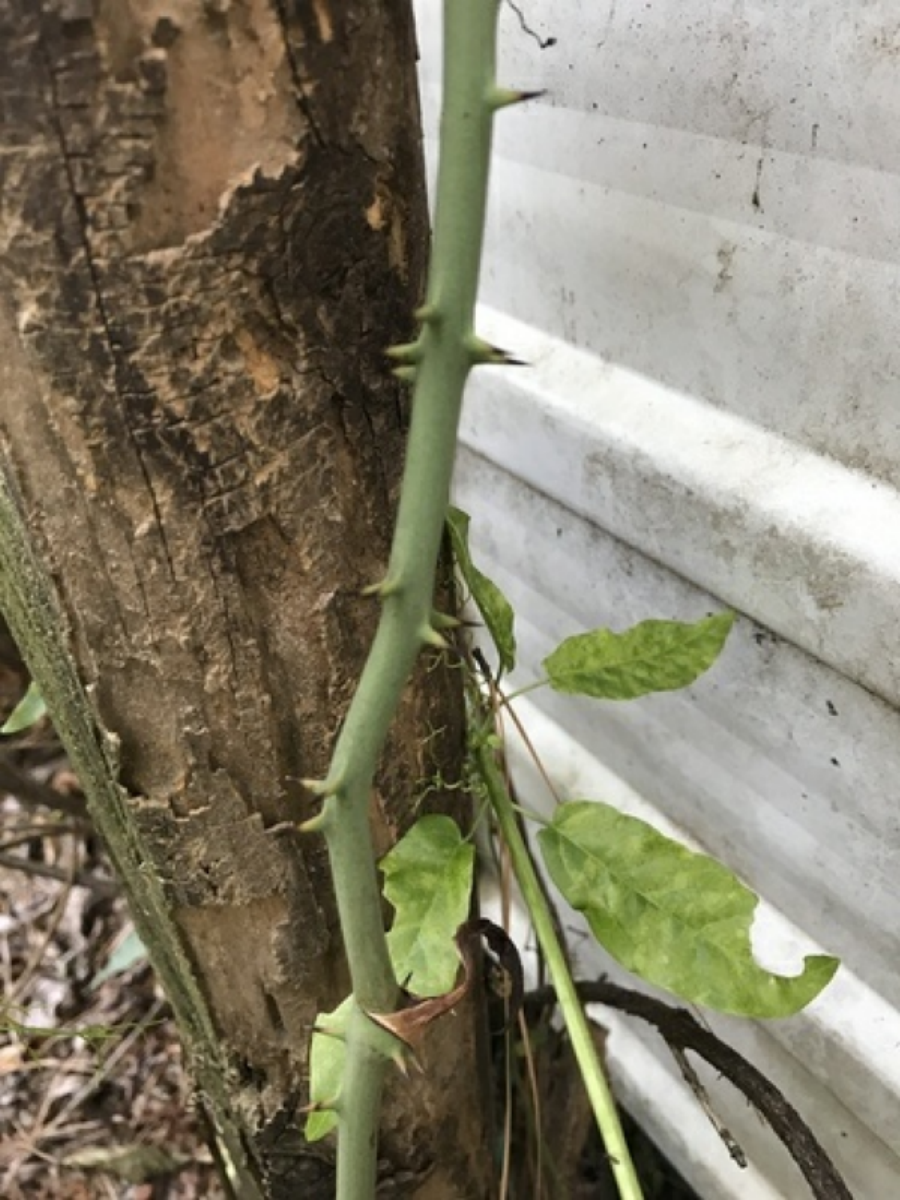} &
\includegraphics[width=0.19\linewidth, height=0.24\linewidth ]{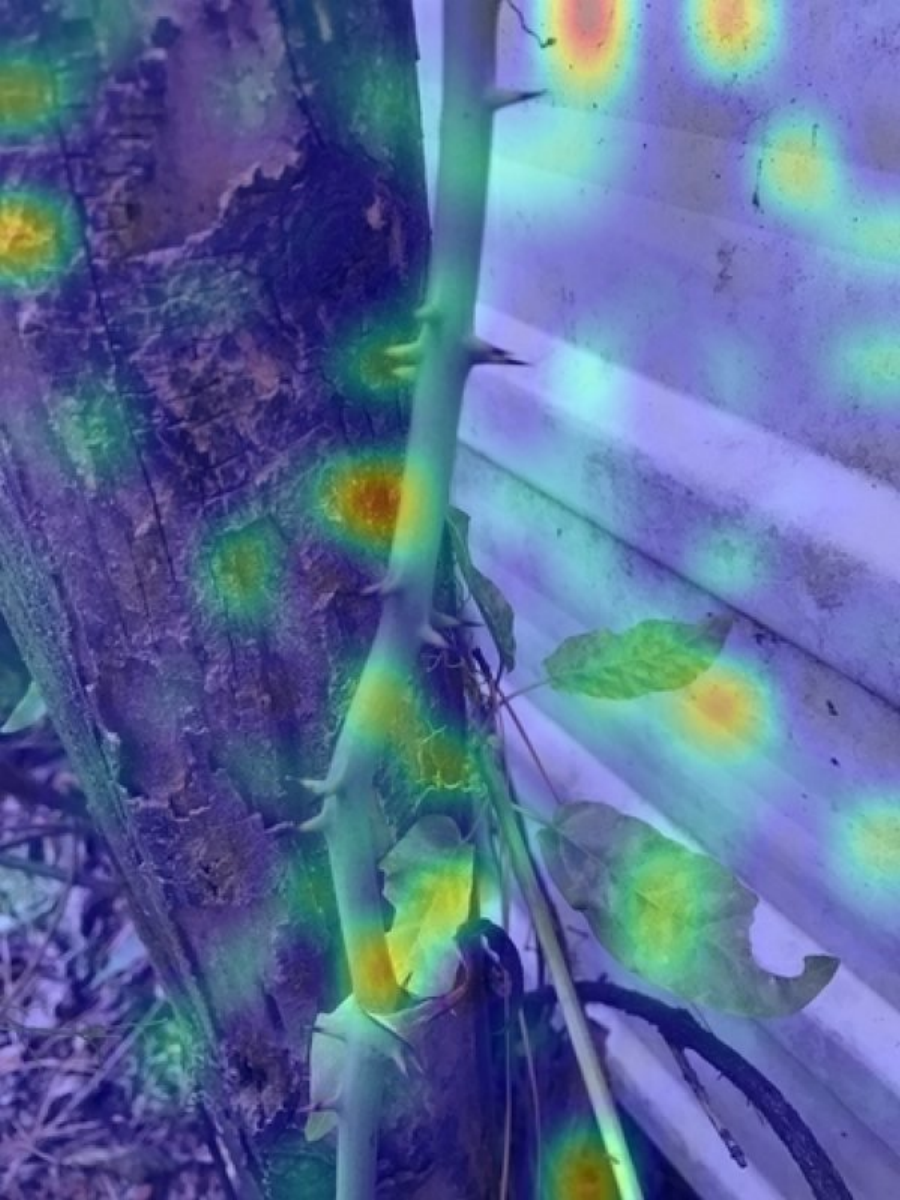} &
\includegraphics[width=0.19\linewidth, height=0.24\linewidth ]{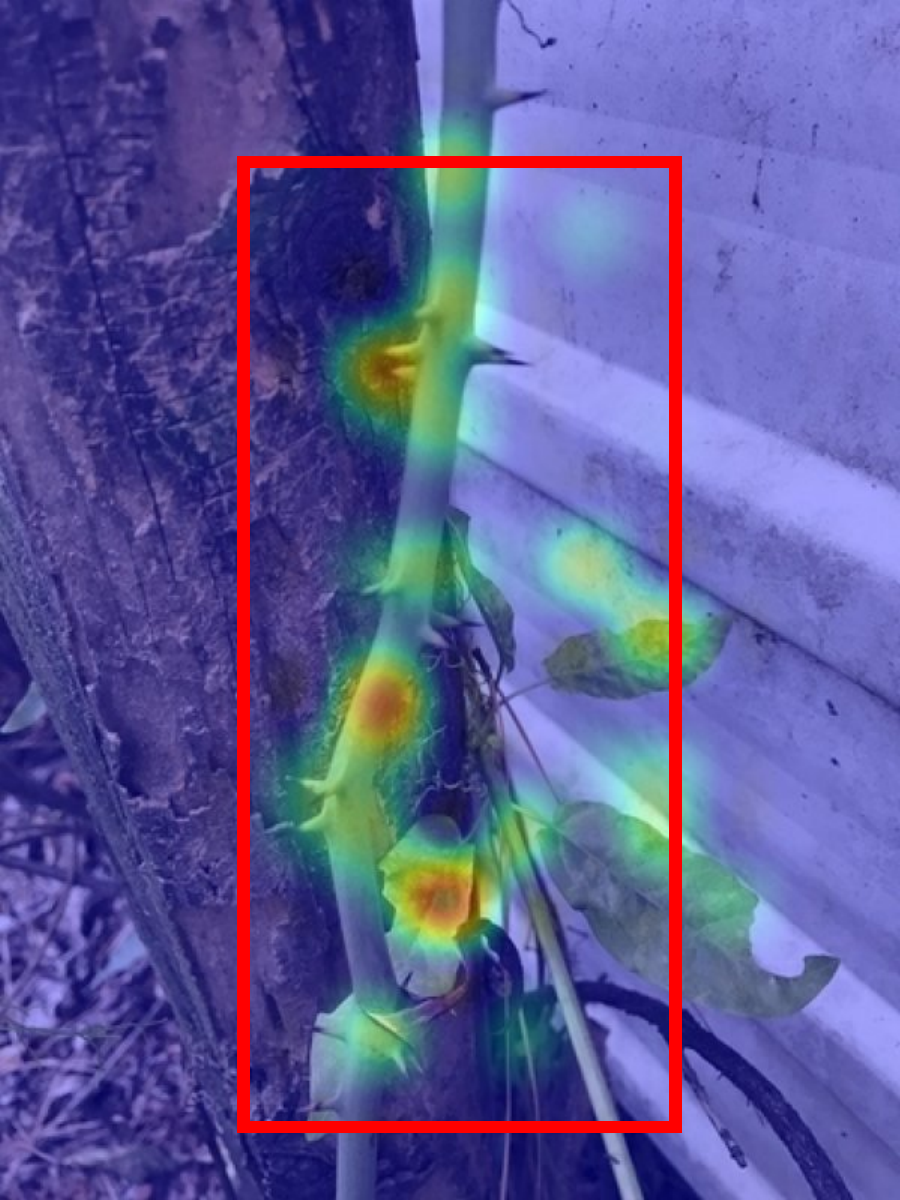} & &
\includegraphics[width=0.19\linewidth, height=0.24\linewidth ]{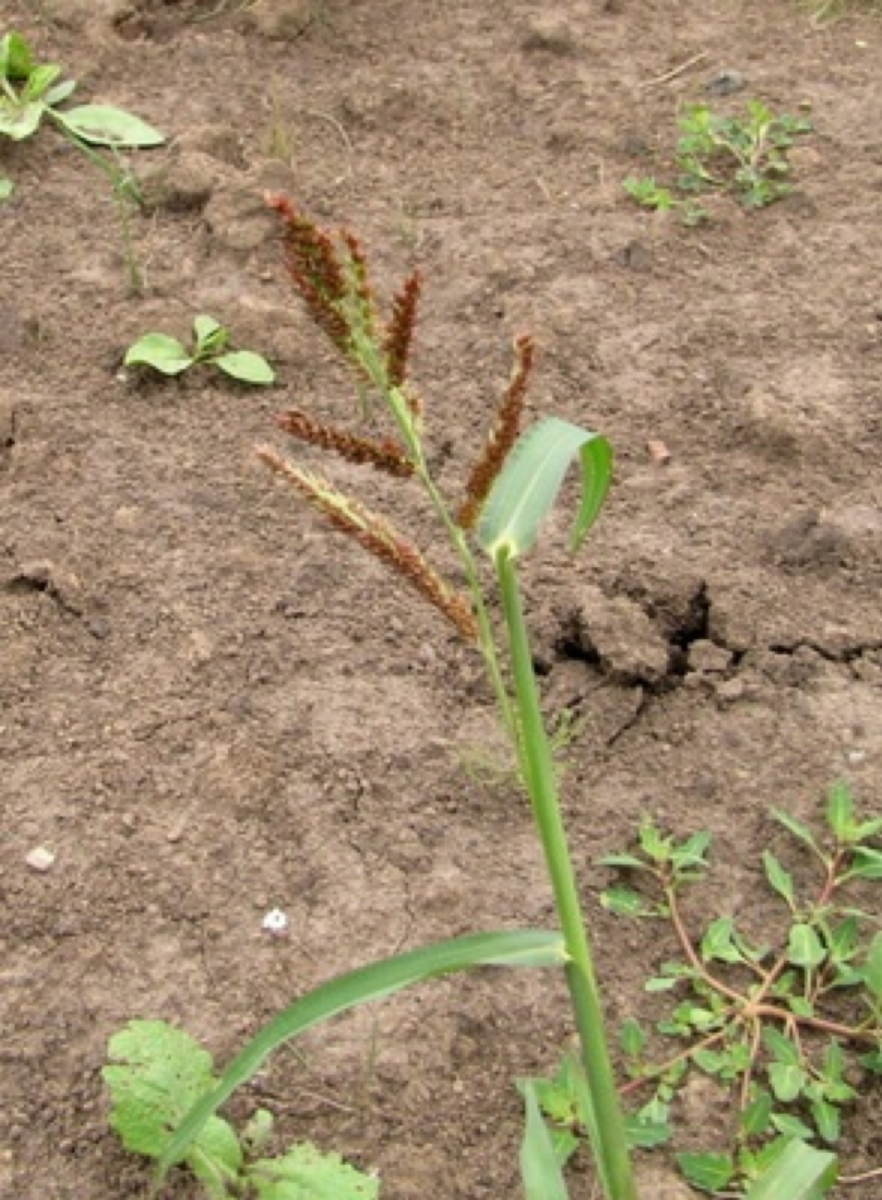} &
\includegraphics[width=0.19\linewidth, height=0.24\linewidth ]{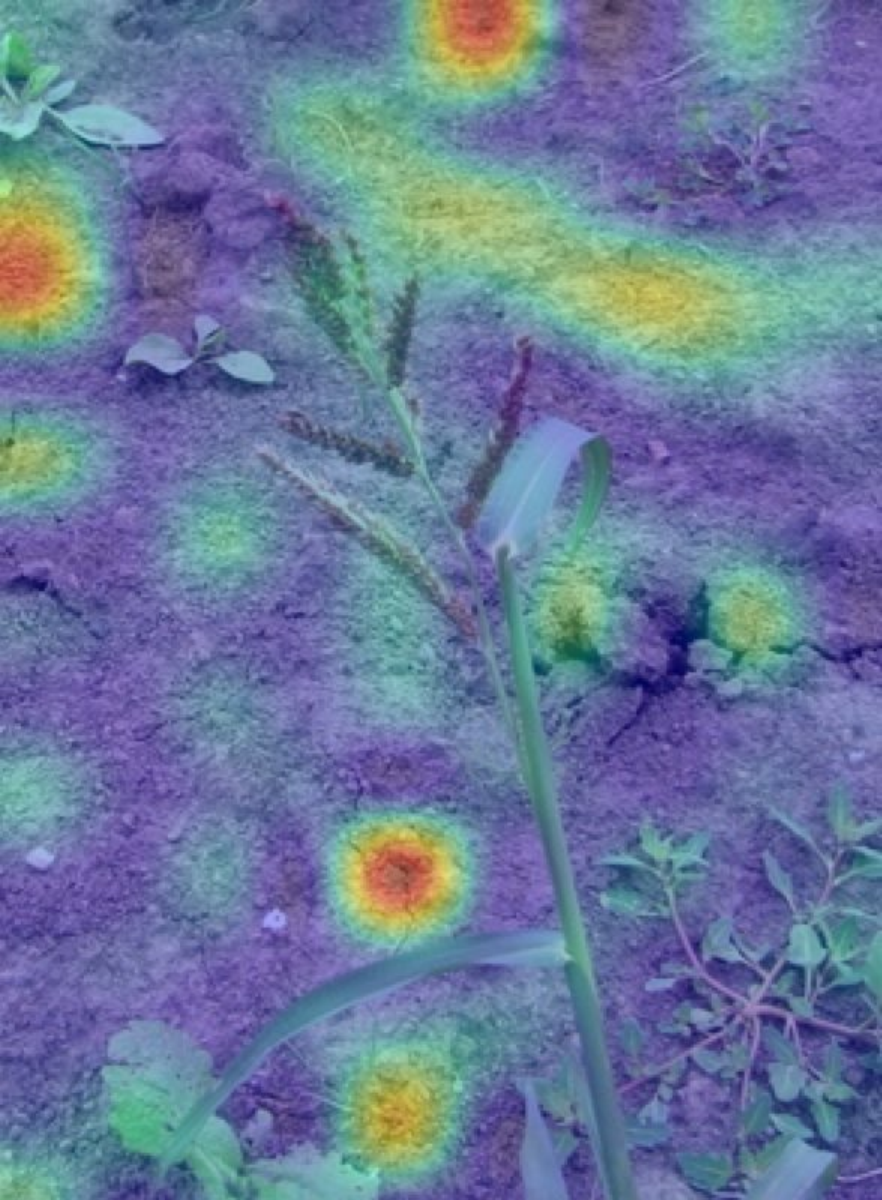} &
\includegraphics[width=0.19\linewidth, height=0.24\linewidth ]{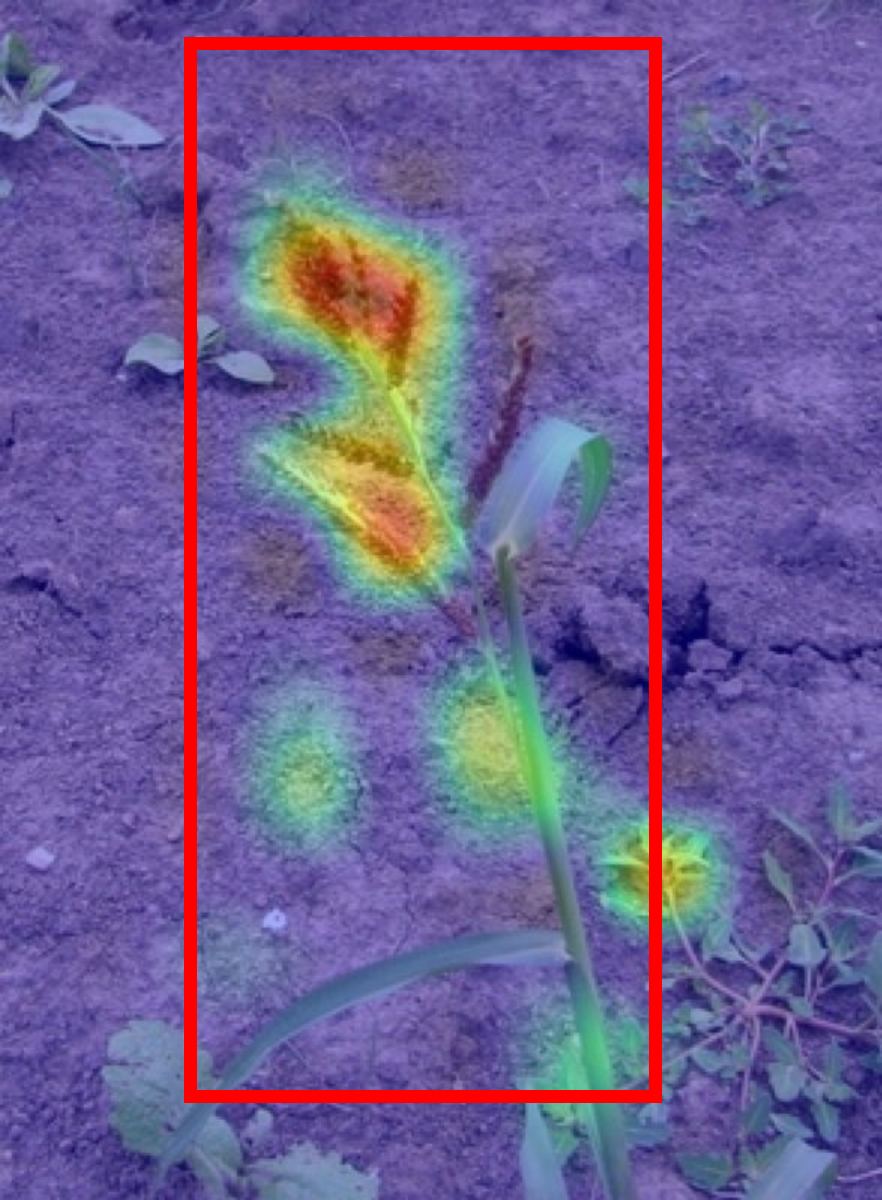} \\

\includegraphics[width=0.19\linewidth, height=0.24\linewidth ]{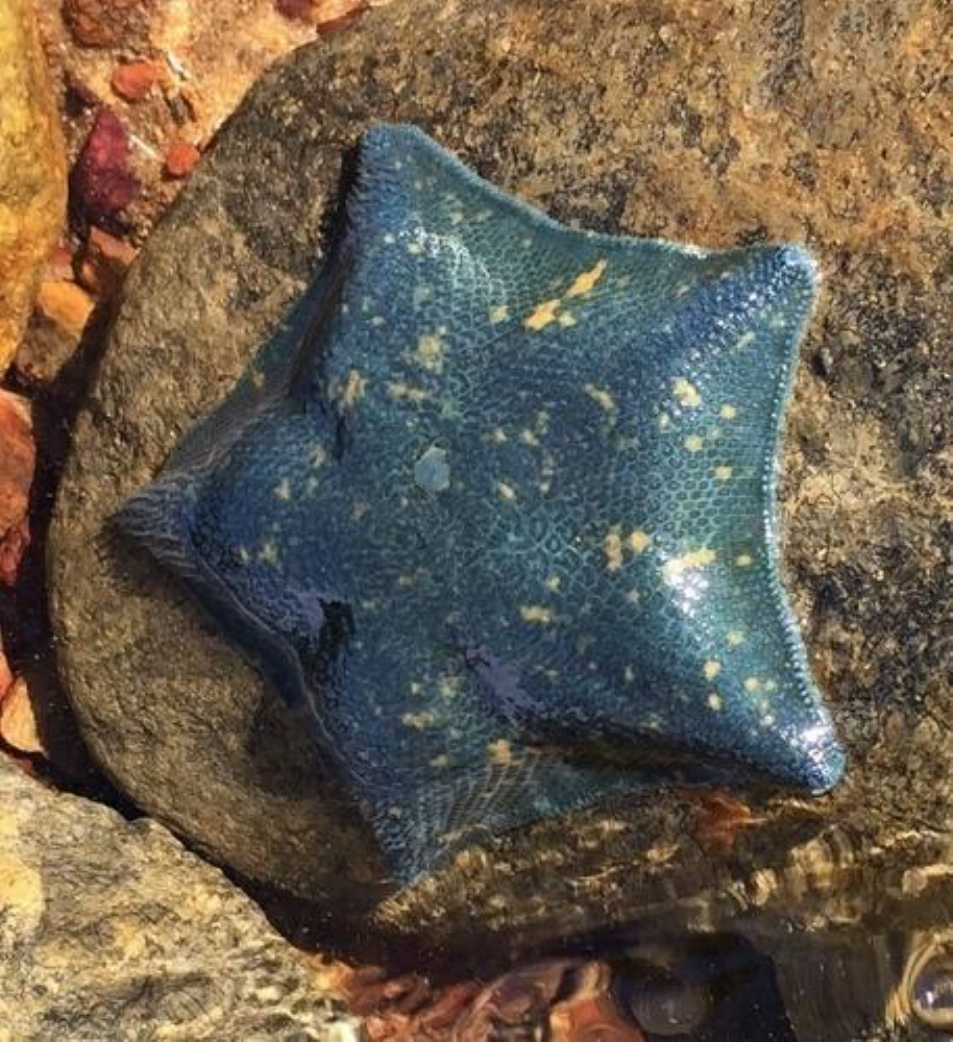} &
\includegraphics[width=0.19\linewidth, height=0.24\linewidth ]{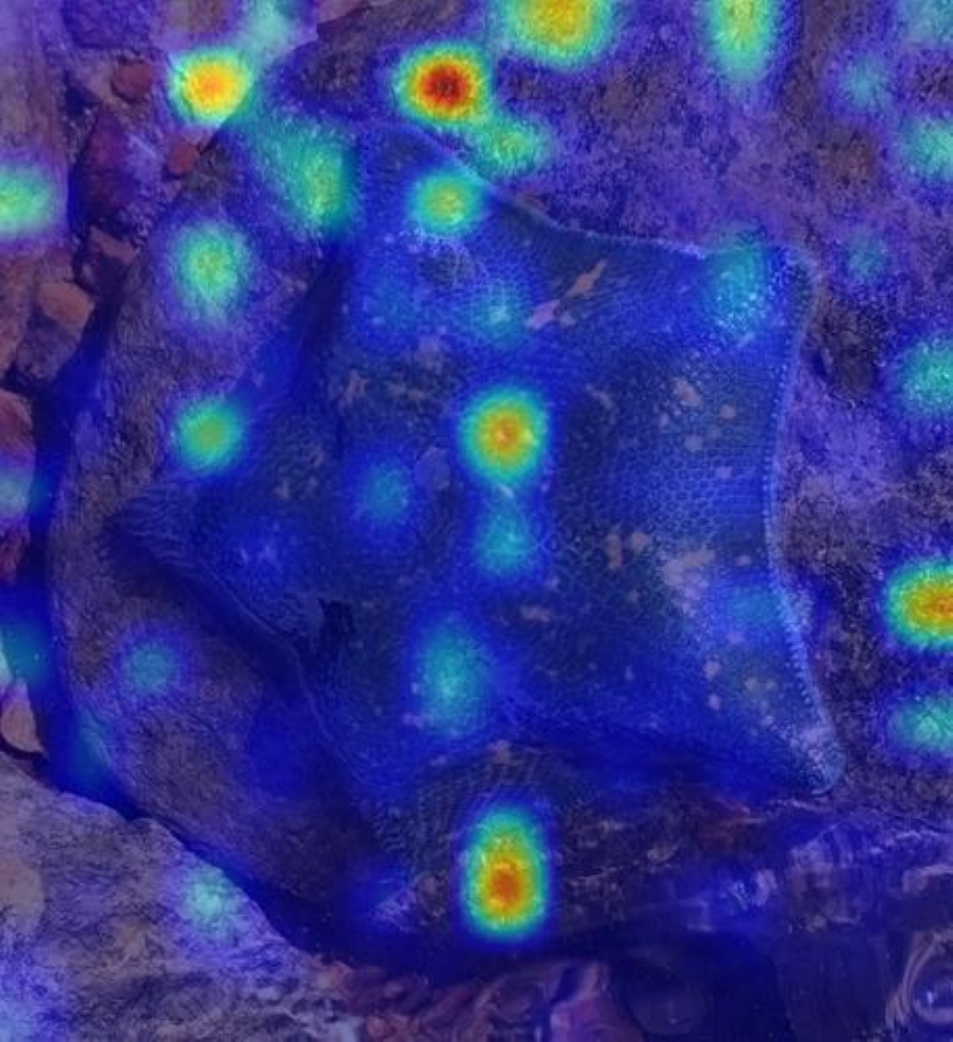} &
\includegraphics[width=0.19\linewidth, height=0.24\linewidth ]{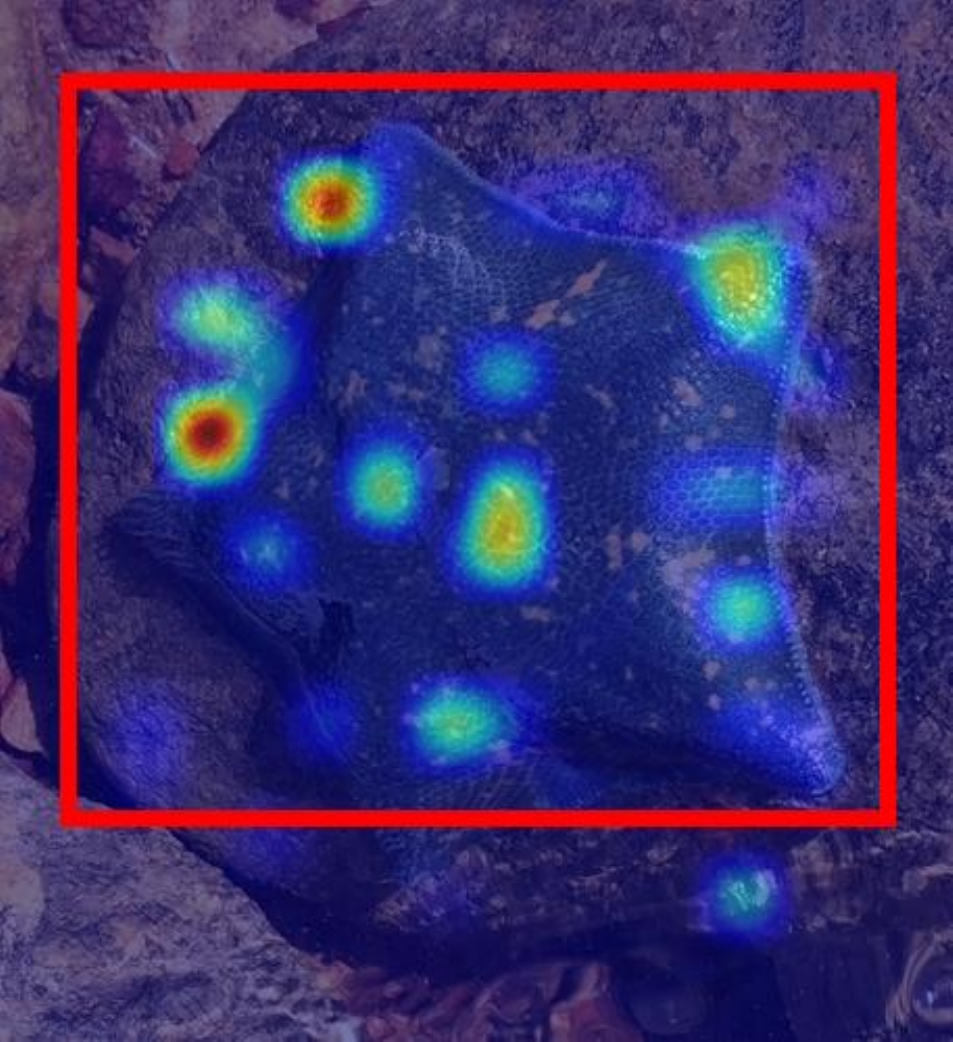} & &
\includegraphics[width=0.19\linewidth,
height=0.24\linewidth
]{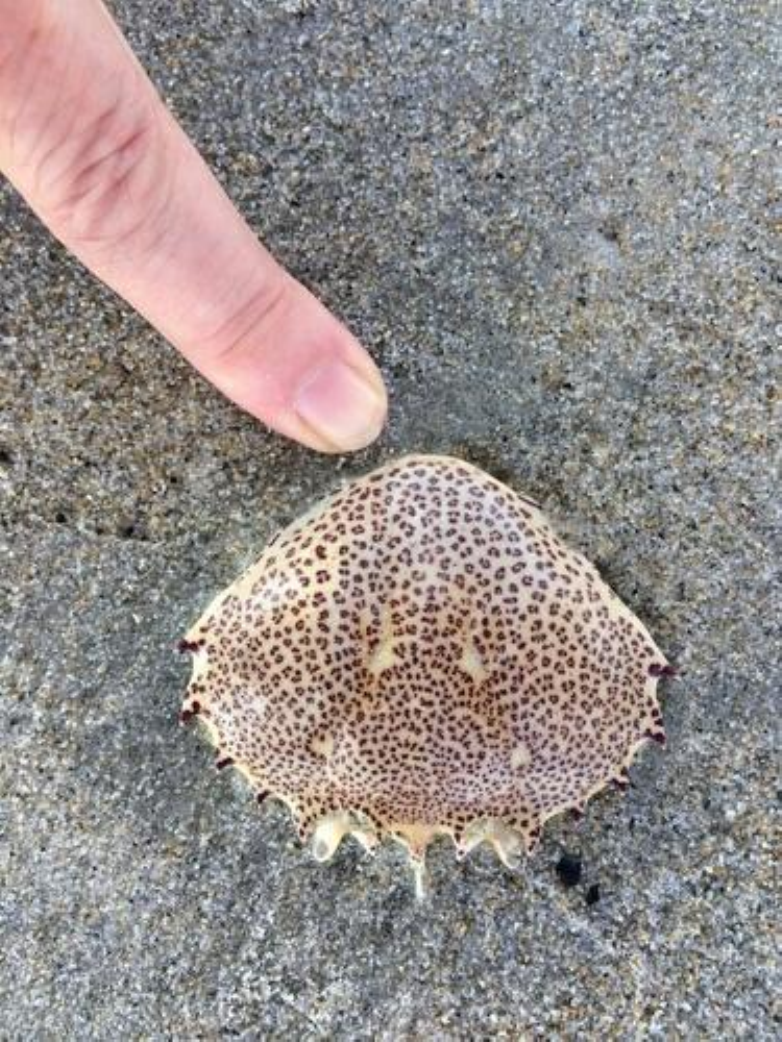} &
\includegraphics[width=0.19\linewidth, height=0.24\linewidth ]{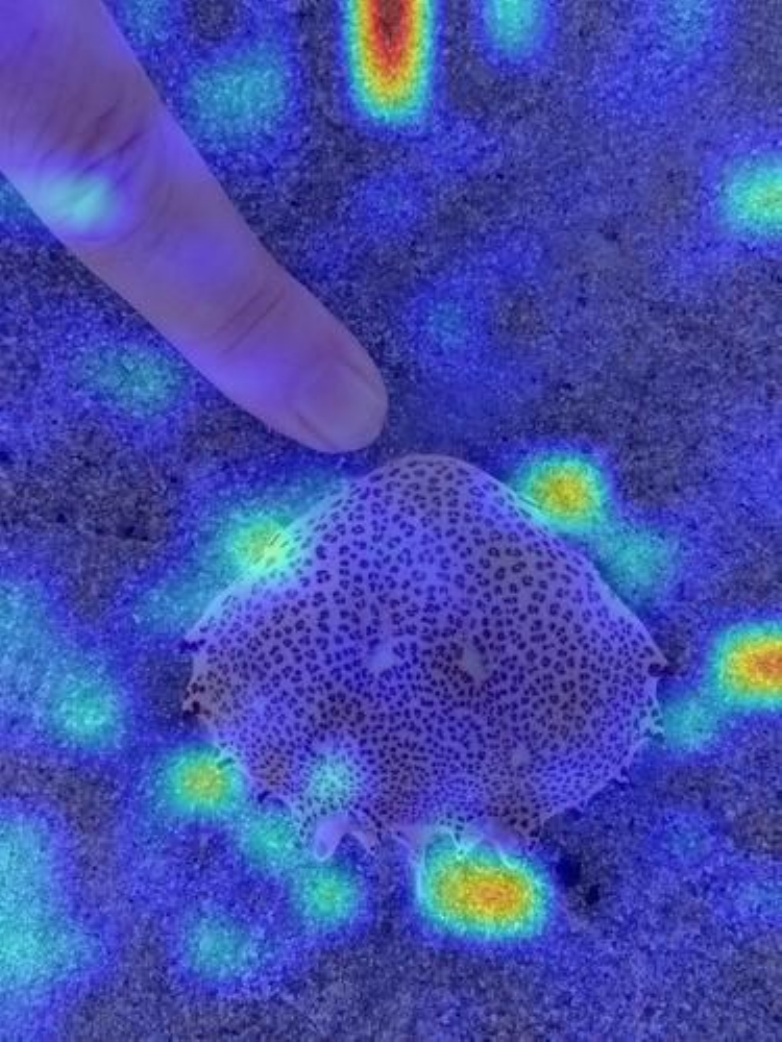} &
\includegraphics[width=0.19\linewidth, height=0.24\linewidth ]{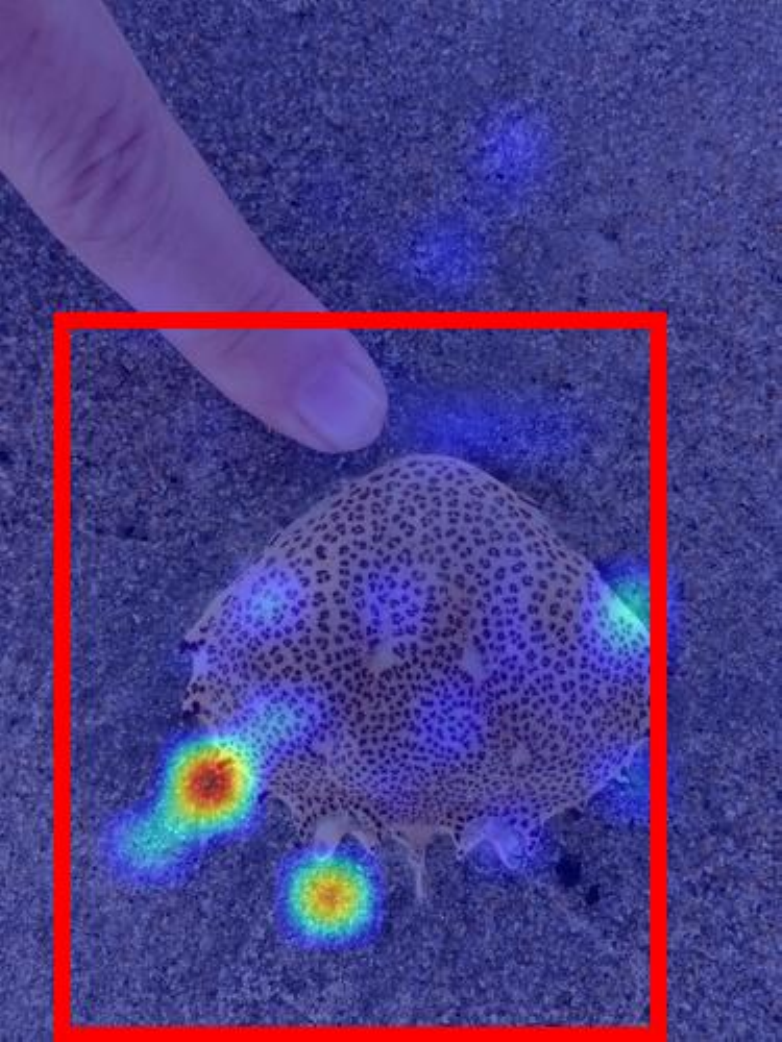} \\

\end{tabular}
}
\vspace{-0.2cm}
\caption{Qualitative examples of attention maps filtering.}
\label{fig:maps}
\end{figure}

\subsection{Visual Evidence Selection}
\tinytit{Attention Map Filtering}
To obtain spatial attention maps, we project the attention scores of visual tokens back onto their corresponding image regions, accounting for each visual tokenization scheme of the model. For the Qwen2-VL, Qwen2.5-VL, and Qwen3-VL checkpoints used in our experiments, the vision encoder merges each $2\times2$ spatial group of neighboring patches into a single visual token, with a spatial merge size of $2$. We therefore reconstruct the post-merging visual grid, whose height and width are those of the original patch grid divided by~$2$.

For InternVL3.5, we exclude the final set of visual tokens corresponding to the global image thumbnail. The remaining patch representations are arranged into their original two-dimensional grid and processed by the pixel-shuffle operation of the model with a downsampling ratio of $0.5$. Consequently, each visual token combines the features of a spatial $2\times2$ group of vision-encoder patches. When dynamic image tiling is used, we reconstruct the attention map independently for each image tile and map it back to the corresponding region of the original image. 

Fig.~\ref{fig:maps} shows some qualitative examples of the effect of attention-sink filtering. The raw attention maps contain scattered activations, often far from the target object, which can lead to inaccurate bounding-box estimation. After filtering, the attention becomes more concentrated around the queried object, producing the more accurate localizations shown by the red bounding boxes.

\tit{Bounding Box Extraction Method}
To extract visual evidence from the model, we experiment with several strategies to aggregate filtered attention maps and derive bounding boxes that localize the relevant visual regions.
We consider three main bounding box extraction methods:
\begin{itemize}[noitemsep, topsep=0pt]
    \item \textbf{Min-Max Coordinates:} This approach considers all non-zero values in the attention map and computes the bounding box by taking the minimum and maximum coordinates along each axis.

    \item \textbf{Morphological Filtering:} The attention map is first thresholded over 0.1 to create a binary map. Morphological closing operations (dilation followed by erosion) are then applied to obtain a connected region, and the bounding box is computed around it.

    \item \textbf{Weighted Centroid:} This method computes the centroid of the attention map weighted by the attention values themselves. The bounding box is then defined around the centroid using the standard deviation of the attention distribution along each axis, scaled by a multiplier to control box size.
\end{itemize}

For this analysis, we focus on E-VQA. Since the dataset does not provide bounding-box annotations for the queried objects, we use predictions from GroundingDINO~\cite{liu2023grounding} (Base), an open-vocabulary object detector, as proxy ground truth. Table~\ref{tab:target_extraction_metrics} reports results in terms of Intersection over Union (IoU), Coverage, Precision, and Center Distance.
As shown, the weighted centroid method consistently achieves the best trade-off between coverage and IoU scores, indicating that it captures the most relevant regions of the attention map while remaining tightly localized around the target object. The min-max method produces overly large and sparse bounding boxes, leading to lower precision, while the morphological approach generates compact regions but occasionally fails to cover all relevant attention peaks. Overall, the weighted centroid method provides the best trade-off between accuracy and compactness, making it the most effective strategy for bounding box extraction.

\begin{table}[t]
\centering
\footnotesize
\setlength{\tabcolsep}{0.2em}
\resizebox{\linewidth}{!}{
\begin{tabular}{lccccc}
\toprule
\textbf{Method} & IoU $\uparrow$ & Coverage $\uparrow$ & Precision $\uparrow$ & Center Distance $\downarrow$ \\
\midrule
Min-Max Coordinates & 0.352 & 0.891 & 0.500 & 0.090 \\
Morphological Filtering & 0.399 & 0.582 & 0.716  & 0.084 \\
Weighted Centroid & 0.487 & 0.814 & 0.626 & 0.071 \\
\bottomrule
\end{tabular}
}
\vspace{-0.15cm}
\caption{Comparison of bounding box extraction methods for target object localization.}
\label{tab:target_extraction_metrics}
     \vspace{-0.2cm}
\end{table}

\begin{table}[t]
    \centering
    \small
    \setlength{\tabcolsep}{0.85em}
    \resizebox{0.65\linewidth}{!}{
       \begin{tabular}{cc c cc}
        \toprule
        $\tau$ & $\beta$ & & IoU $\uparrow$ & Acc@IoU$\geq$0.5 $\uparrow$ \\
        \midrule
        10 & 1 & & 0.267 & 0.023 \\
        10 & 2 & & 0.461 & 0.360 \\
        10 & 3 & & 0.405 & 0.170 \\
        \midrule
        25 & 1 & & 0.317 & 0.045 \\
        \rowcolor{OurColor}
        25 & 2 & & \textbf{0.489} & \textbf{0.441} \\
        25 & 3 & & 0.388 & 0.090 \\
        \midrule
        50 & 1 & & 0.342 & 0.087 \\
        50 & 2 & & 0.484 & 0.408 \\
        50 & 3 & & 0.374 & 0.029 \\
        \midrule
        75 & 1 & & 0.356 & 0.103 \\
        75 & 2 & & 0.481 & 0.399 \\
        75 & 3 & & 0.368 & 0.013 \\
        \bottomrule
    \end{tabular}
    }
    \vspace{-0.15cm}
    \caption{Analysis of bounding-box extraction hyperparameters applied to the filtered attention maps. We report mean IoU and localization accuracy at $\mathrm{IoU} \geq 0.5$.
    }
     \label{tab:bbox_grid_search}
\end{table}

\tit{Bounding Box Extraction Hyperparameters}
On E-VQA, GroundingDINO detections cover approximately $48$\% of the image area, with the median bounding-box ratio falling within the $40$--$50$\% range. In particular, about $53$\% of the detected regions occupy less than half of the image. We therefore focus the hyperparameter analysis on relatively compact entities, for which localization quality is especially critical: tighter boxes enable the resulting crop to isolate the relevant visual evidence while excluding more of the surrounding background.

Furthermore, Table~\ref{tab:bbox_grid_search} reports a grid search on the COCO~\cite{lin2014microsoft} validation split. We restrict the analysis to objects whose ground-truth bounding boxes cover between $25$\% and $50$\% of the image area, as these relatively compact entities provide a more informative test of whether the attention-based procedure can isolate the target from the surrounding scene. Since COCO is an object-detection benchmark and does not provide question-answer pairs, we generate a synthetic query for each evaluated object category that explicitly refers to the target entity. The question and image are then provided to the MLLM, and the bounding box extracted from its attention map is evaluated against the corresponding COCO annotation. Across the tested configurations, a standard-deviation multiplier of $\beta=2$ consistently provides the strongest localization performance. Combined with a sink threshold at the $25$th percentile, it achieves the highest mean IoU and Acc@IoU$\geq0.5$, and is therefore used in all subsequent experiments.

\begin{table}[t]
    \centering
    \small
    \setlength{\tabcolsep}{0.6em}
    \resizebox{0.95\linewidth}{!}{%
    \begin{tabular}{lcccc}
        \toprule
        Configuration
        & IoU
        & Acc@0.5
        & Coverage
        & Precision \\
        \midrule
        w/o sink filtering
        & 0.388
        & 0.083
        & \textbf{0.999}
        & 0.388 \\

        \rowcolor{OurColor}
        w/ sink filtering
        & \textbf{0.489}
        & \textbf{0.441}
        & 0.869
        & \textbf{0.554} \\
        \bottomrule
    \end{tabular}%
    }
    \vspace{-0.15cm}
    \caption{Effect of attention-sink filtering on visual evidence localization.}
    \label{tab:sink_filter_ablation}
    \vspace{-.2cm}
\end{table}

\begin{table}[t]
    \centering
    \small
    \renewcommand{\arraystretch}{0.75}
    \resizebox{0.65\linewidth}{!}{%
    \begin{tabular}{lccc}
        \toprule
        & & \multicolumn{2}{c}{\textbf{E-VQA}} \\
        \cmidrule{3-4}
        \textbf{Layer Range} & & Single-Hop & All \\
        \midrule

        \rowcolor{TitleColor}
        \multicolumn{4}{l}{
            \textit{Text Highlighting}
        } \\

        First half
        & & 21.6
        & 21.3 \\

        Middle half
        & & 30.1
        & 28.4 \\

        \rowcolor{OurColor}
        Last half
        & & \textbf{35.3}
        & \textbf{32.6} \\

        \midrule

       \rowcolor{TitleColor}
        \multicolumn{4}{l}{
            \textit{Visual Highlighting}
        } \\

        First half
        & & 35.2
        & \textbf{32.6} \\

        \rowcolor{OurColor}
        Middle half
        & & \textbf{35.3}
        & \textbf{32.6} \\

        Last half
        & & 34.1
        & 31.9 \\

        \bottomrule
    \end{tabular}
    }
    \vspace{-0.15cm}
    \caption{
    Layer-range ablation on the E-VQA validation split using Qwen2.5-VL-3B. When varying the textual range, visual highlighting is fixed to the middle half; when varying the visual range, textual highlighting is fixed to the last half. 
    }
    \label{tab:layer_range_ablation}
\end{table}

\tit{Effect of Attention-Sink Filtering}
To isolate the contribution of attention-sink filtering, we apply the same weighted-centroid localization procedure with and without removing sink-associated activations. Following the hyperparameter analysis, we evaluate both variants on the COCO validation subset containing objects that occupy $25$--$50\%$ of the image area, while keeping all other localization settings fixed.
As shown in Table~\ref{tab:sink_filter_ablation}, filtering produces tighter and more accurate regions, substantially improving IoU, Acc@IoU$\geq0.5$, and precision, with only a moderate reduction in coverage. Without filtering,
coverage is nearly $100\%$ because attention sinks often occur near the image boundaries, as also shown in Fig.~\ref{fig:maps}. Removing them therefore yields a better overall localization trade-off.

Importantly, the relatively aggressive threshold  $\tau$ only excludes sink tokens from estimating the target location, not from the final crop. Because the crop is defined around the weighted centroid, filtered patches may still be included within the resulting bounding box.

This design is particularly suitable for KB-VQA, where the queried entity is often small relative to the full image. Restricting localization to the strongest object-conditioned activations reduces the influence of residual attention on background regions and yields a more precise centroid. The subsequent box construction, controlled by $\beta=2$, then restores an appropriate spatial margin around this centroid, compensating for the reduced set of contributing tokens and covering the target more completely. Together, these properties explain why the selected sink threshold improves, rather than harms, localization performance in Table~\ref{tab:bbox_grid_search}.

\subsection{Layer-Range Selection}
Prior work indicates that textual evidence is primarily consolidated in deeper decoder layers, whereas visual grounding tends to emerge in intermediate layers~\cite{kang2025your,liu2025selfelicit,jiang2025devils}. Guided by these findings, we set $L_{\mathrm{txt}}$ to the last half of the decoder and $L_{\mathrm{vis}}$ to its middle half. Table~\ref{tab:layer_range_ablation} empirically validates these literature-guided choices. When varying one modality-specific range, the other is kept fixed to its selected configuration. The final half yields the best performance for textual highlighting, while the middle half achieves the highest Single-Hop accuracy and ties the best overall result for visual highlighting.

\section{Additional Details on Computational Analysis}
\ours requires only a single additional token generation step, followed by lightweight post-processing of the extracted attention maps. The entire procedure remains fully zero-shot and requires neither additional training nor model fine-tuning.

Our implementation is designed to avoid large memory footprint by monkey-patching the forward pass. We retain standard FlashAttention (version 2) for all attention computations while restricting the attention matrix computation to a small subset of $Q K^T$ slices,. 
resulting in $(|\mathcal{T}_{\text{obj}}|+1) \times S$ instead of $S \times S$, where $(|\mathcal{T}_{\text{obj}}|+1) \ll S$ with long contexts since $|\mathcal{T}_{\text{obj}}|$ is on average $\sim$2.

\tit{Token Reduction}
To quantify the impact of these operations on E-VQA test split with Qwen2.5-VL-3B, we measure the average number of input and output tokens. \ours substantially reduces the multimodal input used for final answer generation by an average of $2.3$k tokens per sample, corresponding to a $71.93$\% reduction. Specifically, it removes an average of $124.0$ visual tokens ($54.24$\%) and $2.1$k text tokens ($72.90$\%) per sample. Thus, LoT produces a substantially more compact multimodal input, primarily by filtering irrelevant textual context while also reducing the visual token count.

\tit{Latency Breakdown}
We further measure the average per-sample latency of the two stages on E-VQA test split using Qwen2.5-VL-3B. Baseline answer generation requires $1.057$ seconds per sample. \ours spends $0.341$ seconds on multimodal evidence highlighting and $0.866$ seconds on answer generation from the refined input. Therefore, \ours requires $1.207$ seconds per sample, corresponding to an absolute increase of $0.150$ seconds and a $14.1$\% latency overhead.
Notably, generation from the refined input is faster than baseline generation, decreasing from $1.057$ to $0.866$ seconds per sample, an $18.09$\% reduction.

\tit{Computational Cost}
We also estimate the floating-point operations required by each stage on EVQA test split using Qwen2.5-VL-3B. Baseline answer generation requires $19.995$ TFLOPs per sample, whereas generation from the \ours-refined input requires only $3.783$ TFLOPs, corresponding to an $81.08\%$ reduction. The multimodal evidence highlights requires $24.767$ TFLOPs, resulting in a total cost of $28.551$ TFLOPs per sample for the complete pipeline. This represents a $42.79\%$ increase relative to baseline answer generation.

\tit{Run-to-Run Variability}
All evaluations use greedy decoding with temperature set to $0$ and sampling disabled. Thus, no stochastic token sampling is involved, removing the main source of run-to-run variation in generated answers. We therefore report a single evaluation run for each model--dataset configuration rather than averages and standard deviations over repeated runs. Repeating the complete evaluation across all large-scale MLLMs and benchmarks would incur substantial computational cost while providing limited additional information about sampling variability.

\section{Additional Details on KB-VQA Benchmarks}
In the context of KB-VQA, models are required to answer questions by retrieving and leveraging information from an external knowledge base, typically composed of structured or unstructured resources such as Wikipedia. We evaluate our approach on several representative KB-VQA datasets, all of which link images and questions to Wikipedia entities and provide corresponding knowledge bases. 

\tit{Encyclopedic-VQA~\cite{mensink2023encyclopedic}} This
dataset contains 221k question-answer pairs, each associated with up to five images and covering 16.7k fine-grained entities.  
Questions are divided into \textit{single-hop} and \textit{two-hop} types: single-hop questions can be answered using information from a single Wikipedia page, whereas two-hop questions require sequential retrieval across multiple pages. The dataset is divided into training, validation, and test splits comprising 1M, 13.6k, and 5.8k samples. All experiments are conducted on the test split, which includes 4.8k single-hop questions. The dataset also provides an external knowledge base derived from Wikipedia, comprising about 2M pages, each including the article title, textual content, and associated images. In our experiments, we employ the original 2M-page knowledge base provided with the dataset.

\tit{InfoSeek~\cite{chen2023can}} It contains 1.3M image-question-answer triplets associated with about 11k unique Wikipedia entities. The dataset is divided into training, validation, and test splits, comprising roughly 934k, 73k, and 348k samples, respectively. Both validation and test sets include questions about unseen entities.
InfoSeek also provides an external knowledge base covering approximately 6M Wikipedia entities. Following previous works~\cite{caffagni2024wiki,cocchi2025augmenting}, experiments are conducted using a knowledge base of 100k pages and employing 3-shot examples in the prompt to align the response with the dataset format.

\tit{OVEN~\cite{hu2023open}} This dataset is designed to link an image and a textual query to the corresponding Wikipedia entity, covering approximately 6M entities. OVEN includes nearly 5M training examples, along with validation and test splits, as well as a human-verified evaluation set. Importantly, the dataset explicitly distinguishes between \textit{seen} entities (available during training) and \textit{unseen} entities, enabling evaluation of the model ability to generalize to novel concepts. In our experiments, we employ the same knowledge base used in InfoSeek and report results on the query split of the dataset.

\tit{ViQuAE~\cite{lerner2022viquae}} It is a knowledge-based VQA dataset focused on entity-centric questions grounded in Wikipedia. It contains approximately 3.7k question-answer pairs associated with images and entity-centric queries. Each example links an image to a corresponding Wikipedia entity and requires retrieving relevant encyclopedic knowledge to answer the question. In our experiments, we employ the same knowledge base used in Encyclopedic-VQA. In the main results, we report the exact-match metric of the dataset.

\section{Multimodal Inference Prompts with Markers}
After the relevant textual sentences and visual regions have been identified and extracted, we explicitly guide the model to attend to these elements during its response generation process by marking them with special tokens.
For the textual highlighting, sentences selected as relevant are inserted between the tokens: \texttt{\small<START\_IMPORTANT\_TXT>} and \texttt{\small<END\_IMPORTANT\_TXT>}. While in the visual highlighting, the original image is cropped employing the extracted bounding box and inserted between the markers: \texttt{\small<START\_IMPORTANT\_IMG>} and \texttt{\small<END\_IMPORTANT\_IMG>}. These markers do not represent single tokens, they will simply be tokenized into multiple tokens by the tokenizer.
The prompt-box below illustrates the final structure of the user template used at inference time.

\begin{user_template}

\textbf{<START\_IMPORTANT\_IMG>} 

[image\_cropped]

\textbf{<END\_IMPORTANT\_IMG>}

[question]

The following paragraphs may contain useful information to help answer the question correctly:

[sentence]

...

\textbf{<START\_IMPORTANT\_TXT>} 

[evidence\_sentence]

\textbf{<END\_IMPORTANT\_TXT>}

...

[sentence]

\end{user_template}

\smallskip
Finally, the prompt box below presents the system prompt. For each modality, the prompt explicitly instructs the model to prioritize and ground its reasoning in the content enclosed within the corresponding special markers.

\begin{system_prompt}
Answer the encyclopedic question about the given image. Don’t mention the visual content of the image in your output. Directly output the answer of the question according to the context.

If the paragraphs do not contain the information required to answer the question, you should answer the question using your knowledge.

\textbf{<START\_IMPORTANT\_IMG>} and \textbf{<END\_IMPORTANT\_IMG>} are used to mark the important visual evidence. Do not output the markers.

\textbf{<START\_IMPORTANT\_TXT>} and \textbf{<END\_IMPORTANT\_TXT>} are used to mark the important textual evidence. Do not output the markers.

\end{system_prompt}

\section{Additional Qualitative Results}

\begin{figure*}[t]
    \centering
    \includegraphics[width=\linewidth,]{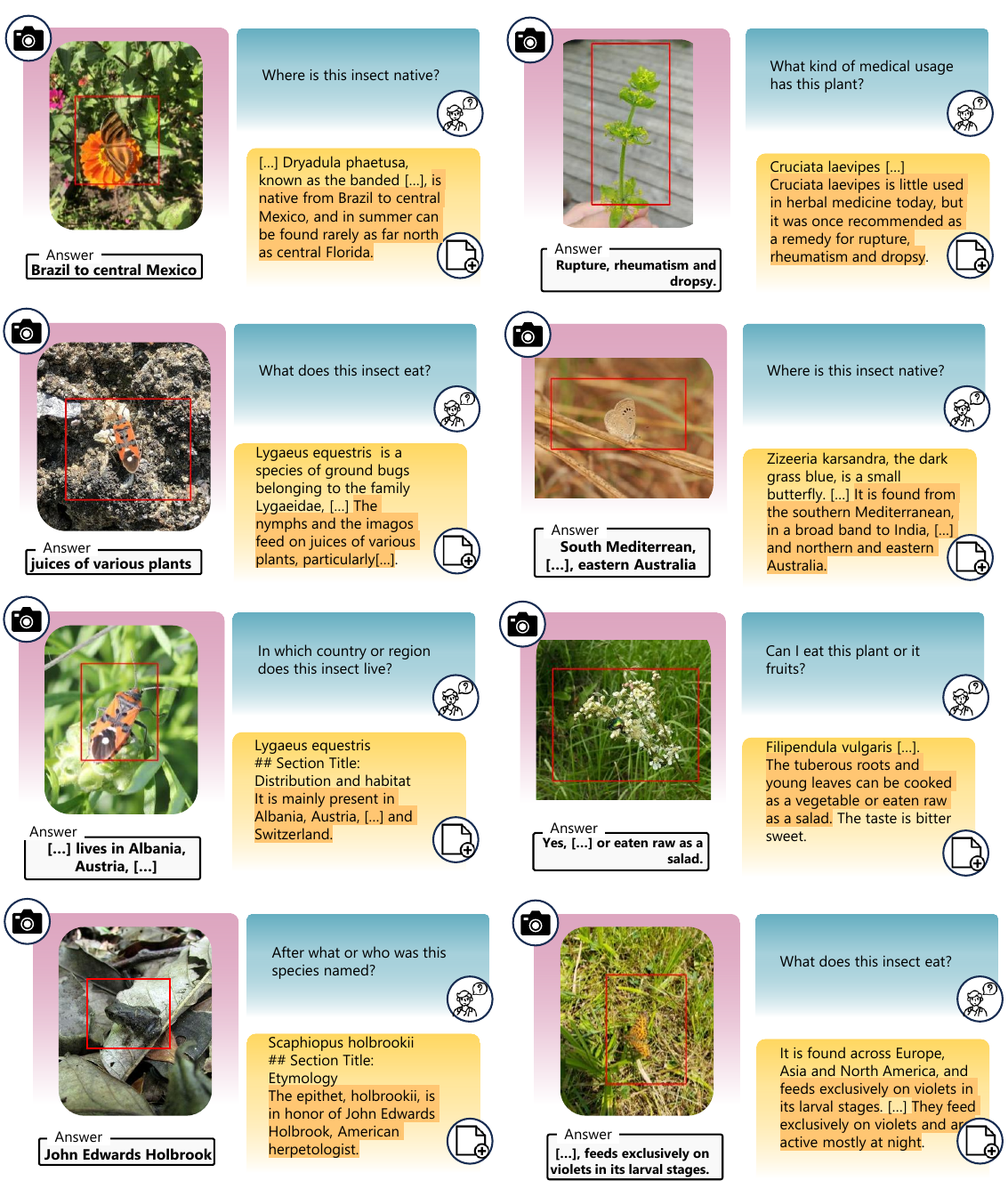}
    \vspace{-0.5cm}
    \caption{Qualitative examples of \ours highlighting query-relevant visual regions and textual evidence, enabling the model to generate the correct answer on E-VQA.}
    \label{fig:qualitatives_supp}
    \vspace{-0.3cm}
\end{figure*}

We also report additional qualitative results in Fig.~\ref{fig:qualitatives_supp}. The examples show how the model identifies the main subject of the question and localizes it in the image (red bbox), allowing it to focus on the most informative visual regions. On the textual side, \ours highlights the sentences within the retrieved documents that are most relevant for answering the question.

\section{Limitations}
\ours relies on a two-stage inference procedure. Although the refined input substantially reduces the cost of final answer generation, the model must still perform an additional forward pass and repeat the prefill stage, increasing overall latency and computational cost. Moreover, \ours{} requires access to internal attention maps and hidden representations, and is therefore not directly applicable to black-box models exposed only through APIs. For syntactically atypical questions, the lightweight spaCy parser may fail to identify the target object, in which case we fall back to the final input token; nevertheless, this remains substantially cheaper than invoking a separate billion-parameter LLM. The implementation also introduces non-negligible CPU-side preprocessing for token-to-context alignment and sentence-span construction, although these operations could be further optimized.
Finally, \ours{} treats internal attention as a heuristic relevance signal rather than a faithful explanation of the model prediction. Consequently, inaccurate relevance estimates may discard useful evidence, particularly when it is diffuse across the image or distributed across multiple textual passages. Performance is also sensitive to the filtering strength, as shown in Table~\ref{tab:alpha_selection}, highlighting the trade-off between removing distracting context and preserving potentially useful evidence.

\end{document}